\pdfoutput=1
\documentclass{article}

\PassOptionsToPackage{numbers}{natbib}
\usepackage[final]{neurips_2024}

\usepackage{booktabs}
\usepackage{subcaption} 
\usepackage{adjustbox}

\usepackage[ruled,vlined,linesnumbered,algo2e]{algorithm2e}
\usepackage{algorithmic}
\usepackage{algorithm}
\usepackage{xcolor}
\usepackage{graphicx}
\definecolor{commentblue}{RGB}{4, 20, 110} 

\def\PMAThresh{\mathbb{\rho}}

\def\E{\mathbb{E}}
\def\de{\overset{\Delta}{=}} 
\def\R{\mathbb{R}}
\def\T{{ \mathrm{\scriptscriptstyle T} }} 

\def\loss{\mathcal{L}}
\def\ntest{n_{\textrm{test}}}
\def\Value{v}
\def\shap{\textsc{shap}}
\def\param{W}
\def\weight{\lambda}
\def\E{\mathbb{E}}

\newcommand{\ipfl}{IP-FL}

\usepackage{amsmath}
\usepackage{amssymb}
\usepackage{mathtools}
\usepackage{amsthm}

\usepackage[utf8]{inputenc} 
\usepackage[T1]{fontenc}    
\usepackage{hyperref}       
\usepackage[capitalize,noabbrev]{cleveref}
\usepackage{url}            
\usepackage{booktabs}       
\usepackage{amsfonts}       
\usepackage{nicefrac}       
\usepackage{microtype}      
\usepackage{xcolor}         
\usepackage{wrapfig}
\theoremstyle{plain}
\newtheorem{theorem}{Theorem}
\newtheorem{thm}[theorem]{Theorem}
\newtheorem{corollary}{Corollary}[theorem]
\newtheorem{proposition}[theorem]{Proposition}

\theoremstyle{definition}

\theoremstyle{remark}
\newtheorem{remark}[theorem]{Remark}

\title{IP-FL: Incentivized and Personalized Federated Learning}

\author{%
  Ahmad Faraz Khan$^{1}$ \\
  \texttt{ahmadfk@vt.edu}
  \And
  Xinran Wang$^{2}$ \\
  \And
  Qi Le$^{2}$ \\
  \And
  Zain ul Abdeen$^{1}$ \\
  \And
  Azal Ahmad Khan$^{2}$ \\
  \And
  Haider Ali$^{1}$ \\
  \And
  Ming Jin$^{1}$ \\
  \And
  Jie Ding$^{2}$ \\
  \And
  Ali R. Butt$^{1}$ \\
  \And
  Ali Anwar$^{2}$ \\
}

\begin{document}

\maketitle

\vspace{-2em}
\centerline{$^{1}$Department of Computer Science, Virginia Tech, USA}
\centerline{$^{2}$Department of Computer Science, University of Minnesota, USA}
\begin{abstract}
\label{Abstract}

Existing incentive solutions for traditional Federated Learning (FL) focus on individual contributions to a single global objective, neglecting the nuances of clustered personalization with multiple cluster-level models and the non-monetary incentives such as personalized model appeal for clients. In this paper, we first propose to treat incentivization and personalization as interrelated challenges and solve them with an incentive mechanism that fosters personalized learning. Additionally, current methods depend on an aggregator for client clustering, which is limited by a lack of access to clients' confidential information due to privacy constraints, leading to inaccurate clustering. To overcome this, we propose direct client involvement, allowing clients to indicate their cluster membership preferences based on data distribution and incentive-driven feedback. Our approach enhances the personalized model appeal for self-aware clients with high-quality data leading to their active and consistent participation. Our evaluation demonstrates significant improvements in test accuracy (8--45\%), personalized model appeal (3--38\%), and participation rates (31--100\%) over existing FL models, including those addressing data heterogeneity and personalization.

\end{abstract}
\vspace{-0.3in}
\section{Introduction}
\label{sec:Introduction}



%
%
%
%
Training high-quality models using traditional distributed machine learning requires massive data transfer from the data sources to a central location, which raises various communication, computation, and privacy challenges. In response, Federated Learning (FL)~\citep{konevcny2016federated,McMahan_FL} has emerged as a solution to train models at the source, reducing privacy issues and addressing the need for high-quality models. However, the success of FL relies on resolving various new challenges related to statistical heterogeneity~\cite{fedprox,marfoq2021federated_multitask,fedFomo,fedALA}, scheduling~\cite{TIFL,lai2021oort,HanIEEECLOUD}, and incentive distribution~\cite{FAIR,incentive_cross_silo,FedFaim}. 
To overcome data heterogeneity challenges, personalized FL (pFL) has emerged as an effective solution to generate separate yet related models~\cite{personalization_survey,towards_personalized_FL,Self,collins2022perfedsi,le2022personalized}. Among pFL techniques, similarity-based approaches that use clustering of clients at the aggregator have gained popularity \cite{Mansour2020ThreeAF, Duan2021FedGroupAF, Ruan2022FedSoftSC, ijcai2022p311, ye2022meta}.

However, existing pFL solutions do not include any incentive mechanism, which is crucial in FL to motivate participants to contribute their data and computation resources. 
Existing incentive mechanisms~\cite{FAIR, TIFF, incentive_TPDS} for traditional FL cannot be applied to pFL techniques because they only consider the performance contribution of clients towards training a single objective. In contrast, clients in pFL contribute towards multiple objectives simultaneously~\cite{ditto,Duan2021FedGroupAF,Tang2021PersonalizedFL,fedFomo,marfoq2021federated_multitask,Ruan2022FedSoftSC}.
Furthermore, traditional incentive solutions only provide monetary benefits and do not consider increasing personalized models' appeal as an incentive for encouraging active and reliable participation of clients. Without incentives, participants may provide low-quality data~\cite{FAIR,incentive_TPDS,FedFaim} or opt-out from participation\footnote{By ``opt-out'' we mean the clients voluntarily leave FL due to the lack of incentivization.}~\cite{reverse_auction}, leading to poorly performing pFL models~\cite{incentive_cross_silo,incentive_TPDS}.
Collaboration fairness~\cite{FIFL,FairnessAwareFL} can also be ensured by appropriately rewarding contributions and accounting for data heterogeneity~\cite{fairFL,FedFaim}.


In addition, since existing pFL techniques assume voluntary and consistent participation from clients, the aggregator controls the client selection and training with insufficient knowledge of clients' training capacity, availability, frequency of new incoming data, clustering preferences, and performance requirements from the trained personalized models.
These factors can directly influence the motivation of self-conscious clients to participate consistently. In this paper, we show that these factors cause frequent opt-outs from uninterested clients due to uninformed clustering decisions by the server and low personalized model appeal (PMA)\footnote{Akin to global model appeal~\cite{cho2023maximizing}, we propose a new metric to measure the personalized model appeal.}, which leads to reduced pFL performance. We also show that solving personalization and incentivization as interrelated challenges yields better outcomes for pFL than solving them as separate problems. Incentives can establish a feedback mechanism by providing a cost-benefit analysis guiding clients to make informed decisions in the pFL process. However, this requires new paradigms for clustered pFL using data distribution information available to clients via their preferences and designing incentive mechanisms for increasing pFL appeal to reduce client opt-outs.
In this paper, we propose {\ipfl} that combines clustering-based pFL with token-based incentivization. Unlike prior works that control clustering from the server side, {\ipfl} allows clients to estimate the importance of each cluster and send their preferences for joining them to the aggregator as bids.  To identify a cluster's importance for a client we use the importance weight of the cluster model as defined by FedSoft~\cite{Ruan2022FedSoftSC}. Clients also use the importance weights for weighted local aggregation during single-shot personalization. This client-driven clustering approach results in accurate clustering because clients can attain a global perspective from their local dataset which is only accessible to them and the importance weights information of each cluster. 
This allows them to make informed decisions that the server cannot make, resulting in improved PMA and reduced opt-outs.
%
%
To incentivize clients for consistent participation, {\ipfl} motivates clients to join clusters with the clients that are most similar to them, maximizing their contribution to the cluster and, in turn, their rewards. Good quality cluster-level models then produce more appealing personalized models for each client. The incentive mechanism treats clients as both providers and consumers. As a consumer, the client tries to attain a certain level of personalized model appeal, so it pays the provider to spend resources to participate in training for the said model in each round. Whereas as a provider, the client earns a profit based on its marginal contribution to training the cluster models. The marginal contribution is calculated with a Shapley Value approximation due to the large computational overhead of the original algorithm~\cite{shapley_overhead,shapley_overhead_4}.

\vspace{-4pt}
\paragraph{Contributions.}
\label{subsec:Contributions}
\vspace{-6pt}



In this work, we propose {\ipfl}, an incentive-driven pFL method. The key contributions of {\ipfl} are as follows: First, it integrates client preferences for personalization and provides a feedback mechanism via contribution-based incentives that results in accurate clustering choices resulting in improved personalization. Second, {\ipfl} introduces novel incentives such as improved personalized models' appeal for clients to prevent opt-outs. Third, {\ipfl} has the added advantage of creating personalized models for unseen clients with unknown data distributions that perform similarly to seen clients without requiring additional training. 
Lastly, we show the efficacy of {\ipfl} with a comprehensive theoretical analysis and rigorous experimental verification, demonstrating its superiority compared to other pFL solutions. In particular, we observe an 8--45\% test accuracy improvement of the cluster models, 3--38\% improvement in personalized model appeal, and 31--100\% increase in the participation rate compared to a wide range of FL and pFL solutions.
\section{Related Work}
\label{subsec:Related work}

\textbf{Cluster-based pFL:} Works like FedSoft~\cite{Ruan2022FedSoftSC}, FedGroup~\cite{Duan2021FedGroupAF}, and~\cite{Tang2021PersonalizedFL} employ clustering techniques in pFL. FedSoft performs soft clustering by matching client data distributions, whereas FedGroup evaluates client gradient similarities through the Euclidean distance of decomposed cosine similarity. ~\cite{Tang2021PersonalizedFL} determines personalization-generalization trade-offs via a bi-level optimization problem. Despite these, issues like clustering overhead and client distribution overlap restrictions exist. Other noteworthy models are IFCA~\cite{IFCA}, focusing on loss-based clustering, and~\cite{Mansour2020ThreeAF}, presenting three personalization strategies.
\textbf{Other pFL models:} Meta-learning techniques for rapid personalized model training are found in works like Per-FedAvg~\cite{fine_tuning} and regularization models~\cite{hanzely2021federated, Hanzely2021PersonalizedFL}. Multi-task learning models include Ditto~\cite{ditto}, FedALA~\cite{fedALA},  and pFedMe~\cite{dinh2022personalized}. Additionally, FedFomo~\cite{fedFomo} introduces adaptive local aggregation for personalization, while FedProx~\cite{fedprox} aims to stabilize FL. However, many of these models face challenges in sustaining long-term client engagement and require additional training or restructuring for new clients.
\textbf{Incentivized FL:} FAIR~\cite{FAIR} and FedFAIM~\cite{FedFaim} focus on quality and fairness-based incentives, respectively. The reputation-based and reverse auction method is highlighted in~\cite{reverse_auction}, and a utility-centric approach in~\cite{incentive_TPDS}. Yet, these methods don't integrate seamlessly with all pFL models.

\textbf{Why existing incentive mechanisms cannot be applied directly to pFL frameworks?}
Standard FL incentives targeting a single global goal~\cite{FAIR,reverse_auction,FedFaim,incentive_TPDS} may not align with the multifaceted objectives in pFL, such as cluster-based~\cite{Duan2021FedGroupAF,Tang2021PersonalizedFL,Ruan2022FedSoftSC} or multi-task models~\cite{ditto,Hanzely2021PersonalizedFL,dinh2022personalized}. {\ipfl} employs clustering for pFL, adjusting cluster memberships every $R$ rounds. It distinguishes itself by establishing cluster boundaries while enhancing shared learning via multiple participation at the client level. Moreover, {\ipfl} promotes consistent client participation using PMA and an incentive mechanism based on Individual Rationality (IR) from game theory~\cite{FAIR,IndividualRationalGTIncentive}.



\begin{wrapfigure}{r}{0.4\linewidth}
\vspace{-3.9em}
  \centering
  \includegraphics[width=1.0\linewidth]{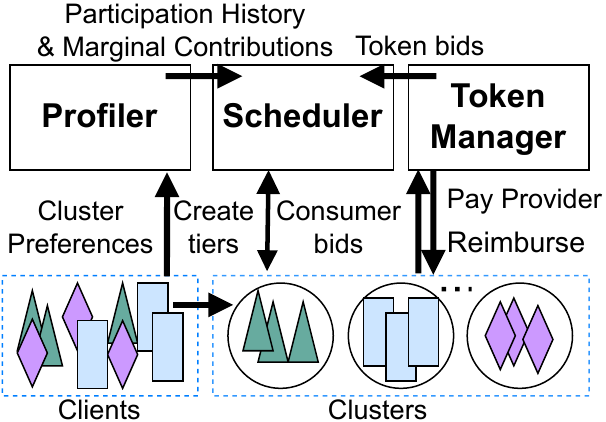}
  \vspace{-15pt}
  \caption{{\ipfl} design}
  \vspace{-1.3em}
  \label{fig:architecture_diagram}
\end{wrapfigure}


\section{Problem Formulation}
\label{sec:Theoretical_Analysis}


In pFL, each client's natural goal is to develop a model that maximizes its test accuracy. Different clients may have varying thresholds for the minimum accuracy gain required to justify their participation in pFL. We denote this self-defined threshold as \(\PMAThresh_{i}\), where \(i \in [N]\). The threshold \(\PMAThresh_{i}\) is defined as the test accuracy that client \(i\) would achieve if it used the conventional FedAvg approach. Therefore, \(\mathrm{PMA}_{i}\) represents the gain in performance from pFL compared to vanilla FL using FedAvg for a client \(i\) among \(N\) clients. PMA is analogous to GMA from~\cite{cho2023maximizing}, however, our analysis and experiments demonstrate that creating a single global model may not align with the interests of all clients. We formalize the concept of PMA and opt-outs as follows:

\vspace{-1em}
\begin{equation}
\label{eq:PMA}
    \mathrm{PMA}_{i} = {f_{i}(w_k) - \PMAThresh_{i}} \mid i \in [N], k \in [K]
\end{equation}
\vspace{-1em}
\begin{equation}
\label{eq:optouts}
    \textrm{opt-outs} = \sum_{i=1}^{N} H(\mathrm{PMA}_{i}) \quad \text{for } i \in [N], \quad H(x) = 
\begin{cases} 
0 & \text{if } x \leq 0 \\
1 & \text{if } x > 0 
\end{cases}
\end{equation}
\vspace{-0.5em}
\vspace{-1em}

Where \(f_{i}(w_k)\) is the test accuracy achieved by pFL for client \(i\) with model parameters \(w_k\).

However, pFL faces the challenge of incentive distribution, as this approach necessitates considering the multidimensional nature of personalization. Furthermore, data-shift and feature-shift in client data over time complicate clients' decisions on whether continued participation is beneficial, leading to potential opt-outs. Therefore, we aim to \textit{develop a pFL framework that provides specialized incentives for personalized training, thereby reducing opt-outs and increasing $\mathrm{PMA}$ to ensure successful collaboration in pFL.} 

\textbf{Theoretical foundation}

We study the following particular case to develop insights. Suppose there are $m$ clients in total, each observing a set of independent Gaussian observations $z_{i,j} \sim \mathcal{N}(\mu_i, \sigma^2), j= 1, \ldots, n_i$, with a personalized task of estimating its unknown mean $\mu \in \mathbb{R}$. 
The quality of the learning result, denoted by $\hat{\mu}$, will be assessed by the mean squared error $\E_i (\hat{\mu} - \mu)^2$, where the expectation $\E_i$ is taken with respect to the distribution of client $i$.
It is conceivable that if clients' underlying parameters $\mu_i$'s are arbitrarily given, personalized FL may not boost the local learning result. To highlight the potential benefit of cluster-based modeling, we suppose that the $m$ clients can be partitioned into two or more subsets, a common assumption in related works~\cite{ditto,IFCA,Ruan2022FedSoftSC}. For simplicity of the problem, we make a case for two subsets: one with $m_1$ clients, say $T_1 =  \{1,\ldots,m_1 \}$, and the other with $m_2$ clients, say $T_2 = \{ m_1+1, \ldots, m\}$, whose underlying parameters are randomly generated as follows: 
\begin{align}
        \mu_i \sim \mathcal{N}(\beta_1, \tau^2) \mid \quad i \in T_1, \hspace{1em}  \mu_i \sim \mathcal{N}(\beta_2, \tau^2) \mid \quad i \in T_2 . 
\end{align}
Here, $\beta_1$ and $\beta_2$ can be treated as the root cause of two underlying clusters. We will study how the values of sample size $n_i$, data variation $\sigma$, within-cluster similarity as quantified by $\tau$, and cross-cluster similarity as quantified by $|\beta_1-\beta_2|$ will influence the gain of a client in personalized learning.
To simplify the discussion, we will assess the learning quality (based on the mean squared error) of any particular client $i$ in the following three procedures: 

\textbf{Local training}: Client $i$ only performs local learning by minimizing the local loss $L_i(\mu) = \sum_{j=1}^{n_i} (\mu - z_{i,j})^2,$ and obtains $\hat{\mu}_i = n_i^{-1}  \sum_{j=1}^{n_i} z_{i,j}$. Thus, the corresponding error variance is\looseness=-1
\begin{align}
    e(\hat{\mu}_i) = \E_i (\hat{\mu}_i - \mu_i)^2 = \frac{\sigma^2}{n_i}. \label{eq1}
\end{align}
\vspace{-1.2em}

\textbf{Federated training}: Suppose the FL converges to the global minimum of the loss, 
$\sum_{i=1}^{m} \frac{n_i}{n} L_i(\mu) , \quad n \de \sum_{i=1}^{m} n_i,$
which can be calculated to be $\hat{\mu}_{\textrm{FL}} = \sum_{i=1}^{m} \frac{n_i}{n} \hat{\mu}_i$.
Consider any particular client $i$. Without loss of generality, suppose it belongs to cluster 1, namely $i \in T_1$. From the client $i$'s angle, conditional on its local $\mu_i$ and assuming a flat prior on $\beta_1$ and $\beta_2$, client $j$'s $\mu_j$ follows $\mu_j \mid \mu_i \sim \mathcal{N}(\mu_i, 2 \tau^2)$ for $j \in T_1$ and $j \neq i$, and $\mu_j \mid \mu_i \sim \mathcal{N}(\mu_i + \beta_2 - \beta_1, 2 \tau^2)$ for $j \in T_2$. 
Then, the corresponding error is
\vspace{-0.5em}
\begin{align}
    &e(\hat{\mu}_{\textrm{FL}}) 
    = \E_i ( \hat{\mu}_{\textrm{FL}} - \mu_i)^2 \nonumber = \biggl\{ \sum_{j \in T_2}  \frac{n_j}{n}(\beta_2 - \beta_1) \biggr\}^2 + \hspace{-1.5em} \sum_{j=1,\ldots,m, j \neq i} \biggl(\frac{n_j}{n}\biggr)^2 \biggl( \frac{\sigma^2}{n_j} + 2 \tau^2 \biggr) + \biggl(\frac{n_i}{n}\biggr)^2 \frac{\sigma^2}{n_i} . \label{eq2}
\end{align}
It can be seen that compared with (\ref{eq1}), the above FL error can be non-vanishing if $\sum_{j \in T_2}  \frac{n_j}{n}(\beta_2 - \beta_1)$ is away from zero, even if sample sizes go to infinity. In other words, in the presence of a significant difference between the two clusters, FL may not bring additional gain compared with local learning.\looseness=-1

\textbf{Cluster-based personalized FL}: Suppose our algorithm allows both clusters to be correctly identified upon convergence. Consider any particular client $i$. Suppose it belongs to Cluster 1 and will use a weighted average of Cluster-specific models.  Specifically, the Cluster 1 model will be the minimum of the loss
$
\sum_{j \in T_1} \frac{n_j}{n_{\textrm{T1}}} L_j(\mu) , \quad n_{\textrm{T1}} \de \sum_{j \in T_1} n_j,
$
which can be calculated to be $\hat{\mu}_{\textrm{T1}} = \sum_{j\in T_1} \frac{n_j}{n_{\textrm{T1}}} \hat{\mu}_i$.
By a similar argument as in the derivation of (\ref{eq2}), we can calculate 
\vspace{-0.5em}
\begin{align}
    e(\hat{\mu}_{\textrm{T1}}) = \sum_{j\in T_1, j \neq i} \biggl(\frac{n_j}{n_{\textrm{T1}}}\biggr)^2 \biggl( \frac{\sigma^2}{n_j} + 2 \tau^2 \biggr) + \biggl(\frac{n_i}{n_{\textrm{T1}}}\biggr)^2 \frac{\sigma^2}{n_i}.
\end{align}
The above value can be smaller than that in (\ref{eq1}). 
To see this, let us suppose the sample sizes $n_i$'s are all equal to, say $n_0$, for simplicity. 
Then, we have
\begin{align} 
    e(\hat{\mu}_{\textrm{T1}}) 
    &= \frac{m_1-1}{m_1^2} \biggl( \frac{\sigma^2}{n_0} + 2 \tau^2 \biggr) + \frac{1}{m^2} \frac{\sigma^2}{n_0} \nonumber = \frac{m_1-1}{m_1^2}  \biggl( \frac{\sigma^2}{n_0} + 2 \tau^2 \biggr) + \frac{1}{m_1^2} \frac{\sigma^2}{n_0} \nonumber =
    \\ & \frac{1}{m_1} \frac{\sigma^2}{n_0} + \frac{m_1-1}{m_1^2} 2 \tau^2,
      \label{eq:within-cluster-bias}
      \text{which is smaller than  (\ref{eq1}) if and only if: } \tau^2 <  \frac{m_1\sigma^2}{2n_0} .
\end{align}
We derive the following intuitions from this analysis: \textbf{R1.} If the within-cluster bias is relatively small, the number of cluster-specific clients is large, and data noise is large, a client will have personalized gain from collaborating with others in the same cluster. 
\textbf{R2.} {\ipfl}'s incentive algorithm rewards accuracy improvement reflected in PMA, which directly correlates with reducing within-cluster bias as per Equation~\ref{eq:within-cluster-bias}.
\textbf{R3.} By association, the incentive algorithm motivates clients to join similar clusters which increases cluster homogeneity and reduces the within-cluster bias. We later show the impact of change in performance with an ablation study of {\ipfl} incentive.




\section{Proposed Methodology}
\label{sec:Design}

In this section, we introduce {\ipfl}, consisting of three key modules: profiler, token manager, and the scheduler (Figure~\ref{fig:architecture_diagram}). The profiler computes and maintains client contributions using Shapley Values approximation (lines 24-27 in Algorithm~\ref{alg:pi-fl-server}), aiding cluster formation. The token manager handles auctions, rewards, and reimbursement transactions (lines 13 \& 14). The scheduler selects clients based on bids and contributions, clustering them for improved homogeneity (lines 20 \& 27-29).
Clients calculate importance weights from cluster models, submit preference bids to the token manager for cluster participation (lines 23-28 in Algorithm~\ref{alg:pi-fl-client}), and generate personalized models (line 29). Clients seek to maximize profits through IR by choosing clusters where they can contribute the most for maximum rewards.

\begin{algorithm}
\footnotesize
   \caption{{\ipfl} (Server)}
   \label{alg:pi-fl-server}
\KwIn{ {\it $R$}: Rounds,
        {\it $K$}: Number of clusters,
        {\it $M_{k}$}: Cluster-level model of cluster $k \in K$,
        {\it $N$}: Number of clients,
        {\it $\zeta_{a}$}: Available Clients,
        {\it $N_{p}$}: Number of clients selected on performance basis,
        {\it $N_{r}$}: Number of clients selected randomly,
        {\it $\zeta_{k}$}: Clients selected for training in cluster $k \in K$,
        {\it sort()}:Timsort~\cite{sort-python}
        
}

  \For{each round $r \in R$}{

    $\zeta_{k} = SelectClients(r)$ for each cluster $k \in K$
    

    \For{cluster $k \in K$}{
    
    Server deploys model $M_{k}$ for training client in $\zeta_{k}$
    
    Token Manager collects bid payments from all willing clients via Eqn. \ref{eq:consumer_cost}

    
    Token manager updates available tokens for round $r$ via Eqn. \ref{eq:update_available_tokens}
    
    
    $U_{k} \gets$ model updates received from clients in $\zeta_{k}$
    
    $M_{k} = FedAvg(U_{k})$~\cite{McMahan_FL}
   }
   
   }
   \vspace{-6pt}
   \SetKwFunction{FLoop}{SelectClients}
    \SetKwProg{Fn}{Function}{}{}
    \BlankLine
    \Fn{\FLoop{$r$}}{
        
        
        

      \uIf{$r = 0$}{
      \For{$k = 1$ {\bfseries to} $K$}{
      $\zeta_{k}^* \gets$ Scheduler selects random clients from $\zeta_{a}$.
      }
      \KwRet {$\zeta_{k}^*$}
      }
      \ElseIf{$r > 1$}{
        \For{$i = 1$ {\bfseries to} $N$}{
        
        ${P_{B}} \gets ClientPreferences(M_{1},...,M_{k}) \mid \forall k \in [K]$ \textcolor{commentblue}{$//$ from Algo.~\ref{alg:pi-fl-client}}
        
        The server calculates marginal contributions $\psi_{{k}{i}}$ of each client within its cluster via Shapley Values approximation in Appendix~\ref{ShapleyValueApprox} Algo.~\ref{alg:ShapleyValues}
        
        $S_{c} = sort(P_{B}, \psi_{{k}{i}})$ \textcolor{commentblue}{$//$Profiler sorts clients by marginal contributions and preference bids}
        
        \For{$k = 1$ {\bfseries to} $K$}{
        $\zeta_{k}^* \gets$ $N_{p}$ clients selected from $S_{c}$ and $N_{r}$ clients randomly from $\zeta_{a}$ by Scheduler.
        }
        }
        \KwRet {$\zeta_{k}^*$}
      }
    }
   \vspace{-5pt}
\end{algorithm}
\subsection{Profiler}
\label{subsec:Profiler}

At the pFL training's onset, the scheduler forms initial clusters by assigning clients randomly. At each round, clients train the cluster-level model on local data and compute the importance weight for each aggregated cluster model $M_{k}$ via Equation \ref{eq:importance_weights}, where $\upsilon_{{i}{k}}$ is the normalized sum of correctly predicted data points on local dataset $D_{i}$. These weights create a personalized model through weighted aggregation via Equation \ref{eq:personalized_model}. Here, $ P_{{i}{k}}$ is client $i$'s personalized model and $\omega_{k}$ is cluster $k$'s weight vector. Through client-centric clustering and participation, clients produce personalized models offline to suit dynamic needs unknown to the server. Clients can also decide on training participation based on budget, past rewards, and importance weights. The clients then bid for the desired cluster in the next training round.

\vspace{-0.8em}
\begin{equation}
\label{eq:importance_weights}
\upsilon_{{i}{k}} = n_{{i}{k}}/n_{k} \in{[0,1]} \mid k \in [K]
\end{equation}
\vspace{-0.9em}
\begin{equation}
\label{eq:personalized_model}
P_{{i}{k}} = \sum_{k=1}^{K} \upsilon_{{i}{k}} \times (\omega_{k})
\end{equation}
\vspace{-1.2em}

Before the start of the next training round, the profiler calculates client marginal contributions using a Shapley Values approximation algorithm given in the Appendix~\ref{ShapleyValueApprox}, providing data quality insights of each client to the scheduler.

\subsection{Token Manager}
\label{subsec:Token Manager}
The token manager coordinates client transactions, operating similarly to a bank executing reward transactions from consumer to provider clients and any reimbursement transactions from providers back to consumer clients. At the beginning of each training round, it conducts an auction for each cluster. Each client $i$ interested in a cluster places its bid using tokens. These bids, represented as $\tau_{p}$, are deducted from the clients' total tokens, $\tau_{i}$, as detailed in Equation \ref{eq:consumer_cost}:

\vspace{-2em}
\begin{equation}
\label{eq:consumer_cost}
\tau_{i} = \tau_{i} - \tau_{p}
\end{equation}

Collected tokens are then added to the Token Manager's overall pool $\tau_{{a}{r}}$ via Equation \ref{eq:update_available_tokens}. $N{p}$ and $N_{r}$ represent the number of clients selected based on performance and random selection respectively.
\vspace{-0.5em}
\begin{equation}
\label{eq:update_available_tokens}
\tau_{{a}{r}} = \tau_{{a}{r}} + (N{p} + N_{r})\times \tau_{p} \mid r \in[1,R]
\end{equation}
The token manager's responsibilities also encompass rewards and penalties as reimbursements. It calculates reimbursements using the utility function, which quantifies the average accuracy improvement or decrement of the cluster model $M_{k}$ over the maximum achieved accuracy in past rounds on the local data of clients in cluster $k$. Then it reimburses consumer clients and penalizes any degradation in the performance of the provider clients to ensure contribution fairness on a finer granularity for consumer and provider clients. This utility and the subsequent reimbursement are defined in Equations \ref{reimbursement_utility} and \ref{reimbursement}:
\vspace{-0.5em}
\begin{equation}
\label{reimbursement_utility}
\begin{aligned}
U_{til} &= \frac{\eta \times (\gamma - \min(\gamma, \max(0.0, \frac{(Acc_{{k}{r}} - Acc_{{k}{max}})}{Acc_{{k}{max}}})))}{\gamma} \mid \eta \in [0,1], \gamma \in [0,1]
\end{aligned}
\end{equation}
\vspace{-0.8em}
\begin{equation}
\label{reimbursement}
\tau_{i} = \tau_{i} - \tau_{{a}{r}} \times U_{til} \mid U_{til} \in [0,\gamma], \forall i \in [N],\forall r \in [1,R]
\end{equation}

$Acc_{{k}{r}}$ is the cluster-level model accuracy in round $r$, $Acc_{{k}{max}}$ is the maximum cluster-level accuracy achieved until this round, $\eta$ and $\gamma$ represent token limitations and accuracy improvement thresholds and $\tau_{{a}{r}}$ is the tokens collected from clients in round $r$. Our approach is inspired by \cite{TIFF}, but it's adapted to account for the impractical assumption of having an IID dataset at the aggregator that perfectly corresponds to the client data distribution within the cluster. Instead, we rely on the local client dataset. Post-reimbursement, the token manager distributes rewards. Provider clients are ranked based on their contributions and participation, and rewards are allocated via Equation \ref{eq:reward_distribution}:

\vspace{-0.8em}
\begin{equation}
\label{eq:reward_distribution}
\begin{aligned}
\tau_{i} &= \tau_{i} + \text{sort}(\psi_{{k}{i}}, \Omega_{{k}{i}}) \times \frac{\tau_{{a}{r}}}{N_{r} \times \frac{(N_{r} + 1)}{2}}  \mid \forall k \in [K], \forall i \in [N], \forall r \in [R]
\end{aligned}
\end{equation}
\vspace{-0.8em}

In Equation~\ref{eq:reward_distribution}, $\psi_{{k}{i}}$ represents client contributions, and $\Omega_{{k}{i}}$ tracks client participation. $\beta$ normalizes token distribution based on provider ranks $\alpha$ in Equation \ref{eq:reward_distribution}. $\tau_{i}$ are client-owned tokens, and $\tau_{{a}{r}}$ are tokens available for distribution by the Token Manager. Through reimbursements to consumers and payments to providers, {\ipfl} incentivizes personalized learning, improving PMA, reducing opt-outs, and ensuring fair client incentives based on contributions to enhance personalized learning outcomes.

\begin{algorithm}
\footnotesize
   \caption{{\ipfl} (Client)}
   \label{alg:pi-fl-client}
    \KwIn{{\it $T_{h}$}: Importance weight threshold,
    {\it $K$}: Number of clusters,
    {\it $M_{k}$}: Cluster-level model of cluster $k$,
    {\it $D$}: Local dataset of client
    }
    \SetKwFunction{FClientPreferences}{ClientPreferences}
    \SetKwProg{Fn}{Function}{}{}
    \BlankLine
    \Fn{\FClientPreferences{$M_{1},...,M_{k}$}}{
    \For{ each cluster $k \in K$}{

    \For {each data point {$d \in D$}}{
    
     The client $i$ computes $\upsilon_{ik}$ importance weight $T_{h}$ of $M_{k}$ model for each data point $d$ via Eqn. \ref{eq:importance_weights}
    }

    \If{$\upsilon_{k} > T_{h}$}{
    Client includes cluster $k$ in their preference list ${P_{B}}$
    
    }
    }
    
    The client generates personalized model $P_{{i}{k}}$ via Eqn. \ref{eq:personalized_model}
    
    \KwRet{${P_{B}}$}
    }
    \vspace{-5pt}
\end{algorithm}

\vspace{-1.5em}
\subsection{Scheduler}
\label{subsec:Scheduler}

The scheduler employs the $SelectClients(r)$ function detailed in Algorithm \ref{alg:pi-fl-server} to choose clients for each round $r$. It collects preference bids ${P_{B}}$ from the token manager and marginal contributions $\psi_{{k}{i}}$ from the profiler for every client $i \in N$ in cluster $k \in K$. Here, $N$ and $K$ denote the total counts of clients and clusters, respectively. With this data, the scheduler groups clients by similar preference bids, arranges them by their marginal contributions, and subsequently picks $N_{p}$ from this ordered list and $N_{r}$ at random. These quantities, $N_{p}$ and $N_{r}$, are adjustable parameters. Following the methodology in~\cite{lai2021oort}, we prioritize exploration in early rounds, initializing $N_r$ to $20\%$ of all clients and decrementing it based on the count of remaining unexplored clients. The reduction strategy, which caps the minimum $N_r$ at $5\%$, guarantees an effective exploration technique. Random selection of $N_{r}$ clients mitigates accuracy bias, echoing techniques from prior studies \cite{McMahan_FL, FLatScale, TIFF, Adaptive_aggregation}. 
By clustering clients with similar preferences, the scheduler reduces within-cluster bias and enhances within-cluster uniformity, culminating in a cluster model that truly mirrors its constituent clients. Section~\ref{sec:Theoretical_Analysis} delves into the significance of this in boosting the PMA.



\section{Convergence Analysis of the {\ipfl}}
\label{ConvergenceAnalysis}
This section explores the convergence properties of {\ipfl}, showcasing its robustness and efficiency. Beginning with foundational propositions, we establish the basis for cluster formation in {\ipfl}. Subsequent theorems then delve into the convergence of personalized and global models across various data distributions, highlighting {\ipfl}'s adaptability for achieving stable and optimal parameters. Proofs for all propositions and theorems are detailed in appendix~\ref{proves}.

\begin{proposition}\label{prop3}
    Let $\{\mathcal{C}_k\}_{k=1}^K$ represent the set of clusters formed by {\ipfl} algorithm, where each cluster $\mathcal{C}_k$ contains clients $\{i_1, \ldots, i_N\}$ with their respective data distributions $\{D_{i_1}, \ldots, D_{i_N}\}$. Cluster will converge and within each cluster $\mathcal{C}_k$, the data distributions of the clients are statistically similar to each other up to a statistical threshold $\delta$, and the within-cluster bias is reduced.
\end{proposition}
\begin{remark}
    Proposition~\ref{prop3} establishes the fundamental concept of cluster formation within {\ipfl}. The result ensures that data distributions within each cluster are homogeneous. Theorem~\ref{theorem5} builds directly on the insights from proposition \ref{prop3}, detailing the convergence of personalized models within clusters which underscores the algorithm's adaptability and efficiency in handling diverse data distributions.
\end{remark}

\begin{thm}\label{theorem5}
    Let $\{\mathcal{C}_k\}_{k=1}^{K}$ be a set of clusters formed by the {\ipfl} algorithm, where each cluster's data follows a Gaussian distribution. If the loss function $L(\theta)$ is convex for the model parameters $\theta$, then the {\ipfl} algorithm converges to a set of stable parameters $\theta^*$ for the global model. Furthermore, once clustering convergence is established, each cluster converges to a set of stable parameters $\theta_k^*$.
\end{thm}

\begin{thm} \label{theorem4}
 Under the assumptions Proposition~\ref{prop3} and Theorem~\ref{theorem5}, there exists a finite number of training iterations $T$ such that for all $ t \geq T$, for every $\epsilon$, the performance metric $\mathrm{PMA}_i$ for each client $i$ within $\epsilon$, i.e., $|\mathrm{PMA}^{(t)}_{i}-\mathrm{PMA}_{i}^{*} |< \epsilon$, where $\mathrm{PMA}_{i}^{*}$ is the optimal.  

 \begin{remark}
As $\mathrm{PMA}_{i}:=f_i(w_k)-\rho_i$, above convergence Theorem~\ref{theorem4} ensures that over the course of training $w_k$ is optimized, and $f_{i}(w_k) $ converges to maximum achievable performance for client $i$. This leads to $\mathrm{PMA}_{i}>0$, which implies the opt-out likelihood is minimized. The following Theorem~\ref{theorem6} demonstrates convergence with general convex and smooth loss functions. This theorem encapsulates the {\ipfl} algorithm's robustness, showcasing its potential to achieve global optimization effectively.
 \end{remark}
 
\end{thm}

\begin{thm}[Convergence of {\ipfl} Algorithm] \label{theorem6}
    Let $ \{c_i\}_{i=1}^N $ be a set of clients participating in the {\ipfl} framework with local loss functions $ \{L_i\}_{i=1}^N $, where each $ L_i $ is convex and $\beta$-smooth. Suppose that the {\ipfl} algorithm updates the global model $M $ by a weighted aggregation of locally updated models using a personalized diminishing learning rate $\eta_t$. If the series \( \sum_{t=1}^{\infty} \eta_t = \infty \) and \( \sum_{t=1}^{\infty} \eta_t^2 < \infty \), then the sequence of global models \( \{M_t\} \) generated by the {\ipfl} algorithm converges to a global minimizer \( M^* \) of the weighted average loss function \( L \), defined as \( L(M) = \sum_{i=1}^N w_i L_i(M) \) where \( w_i \) is the weight corresponding to client \( i \)'s data contribution.
\end{thm}
\begin{table*}[t]
\begin{minipage}{0.38\linewidth}
\caption{Test accuracy (CIFAR10)}
\fontsize{20}{10}\selectfont 
\vspace{-0.23cm}
\centering
\begin{adjustbox}{width=1\linewidth}
\begin{tabular}{lcccccccc}
\toprule
& \multicolumn{4}{c}{10:90} & \multicolumn{4}{c}{30:70} \\
\addlinespace[6pt] 
& \multicolumn{2}{c}{IP-FL} & \multicolumn{2}{c}{FedSoft} & \multicolumn{2}{c}{IP-FL} & \multicolumn{2}{c}{FedSoft} \\
\addlinespace[6pt] 
& c0 & c1 & c0 & c1 & c0 & c1 & c0 & c1\\ 
\addlinespace[6pt] 
\midrule
\addlinespace[15pt] 
$\theta_{0}$ & \textbf{63.7} & 41.3 & 48.9 & 49.5 & 58 & 57.7 & 48 & 48.4\\
\addlinespace[6pt] 
$\theta_{1}$ & 43.7 & \textbf{63.8} & \textbf{50.7} & \textbf{49.6} & \textbf{58.6} & \textbf{58.8} & \textbf{50} & \textbf{50}\\
\addlinespace[5pt] 
\bottomrule
\end{tabular}
\end{adjustbox}
\label{tab:synthetic_PI_FL_and_FedSoft}
\end{minipage}%
\hspace{2pt}
\begin{minipage}{0.61\linewidth}
\centering
\caption{Clusters models accuracy (Synthetic CIFAR10)}
\vspace{-0.23cm}
\fontsize{20}{18}\selectfont 
\begin{adjustbox}{width=1\linewidth}
\begin{tabular}{lcccccccc}
\toprule
& \multicolumn{2}{c}{10:90} & \multicolumn{2}{c}{30:70} & \multicolumn{2}{c}{Linear} & \multicolumn{2}{c}{Random} \\
& c0 & c1 & c0 & c1 & c0 & c1 & c0 & c1 \\
\midrule
\addlinespace[7pt] 
\textbf{\ipfl} & \textbf{62.7(0)} & \textbf{71(1)} & \textbf{53.3(0)} & \textbf{61.9(1)} & 
\vspace{5pt}
\textbf{61.7(0)} & \textbf{59.4(1)} & \textbf{66.9(0)} & \textbf{58.4(1)} \\
\vspace{5pt}
\textbf{FedSoft} & 32.5(0) & 38.6(1) & 20.3(0) & 23.6(0) & 34.42(1) & 49.6(1) & 21.6(1) & 33.12(1) \\
\vspace{5pt}
\textbf{IFCA} & 30.5(0) & 36.2(0) & 18.8(0) & 20.4(0) & 31.2(1) & 46.3(1) & 18.7(1) & 31.24(1) \\
\textbf{FedEM} & 29.7(0) & 37.6(0) & 17.3(0) & 20.6(0) & 30.6(1) & 45.8(1) & 18.4(1) & 30.4(1) \\
\bottomrule
\end{tabular}
\end{adjustbox}
\label{tab:Synthetic_CIFAR10_tier_model_acc_clustering_algos}
\end{minipage}
\vspace{-0.5em}
 
\end{table*}

\section{Experimental Study}
\label{sec:Experimental_Studies}
\subsection{Experimental Setup}
\label{subsec:Experimental_Setup}
We use NVIDIA RTX 3070 GPUs to evaluate {\ipfl} and other pFL methods across four datasets. Our CNN model (32x64x64 convolutional with 3136x128 linear layer parameters) was designed for efficient training on client devices with limited resources, aligning with Cross-Device FL settings~\cite{kairouz2019advances}.

\textbf{CIFAR10 Data:} We utilized the CIFAR10 dataset from FedSoft~\cite{Ruan2022FedSoftSC}, comprising 32 × 32 × 3 images and 10 output classes. We replicated data heterogeneity conditions: 10:90, 30:70, linear, and random partitions. Data was divided into two clusters, $D_{A}$ and $D_{B}$. In the 10:90 partition, 50 clients had $90\%$ training data from $D_{A}$ and $10\%$ from $D_{B}$, while the other $50$ have $10\%$ training data from $D_{A}$ and $90\%$ from $D_{B}$. The 30:70 partition had a $30\%$ and $70\%$ distribution.
\textbf{EMNIST Data:} This dataset contained 28 x 28 images and 52 output classes (26 lowercase and 26 uppercase letters). We employed 10:90 and 30:70 data partitions, along with linear and random partitions. In the linear partition, client $k$ had $(0.5 + k)\%$ training and testing data from $D_{A}$ and $(99.5-k)\%$ from $D_{B}$. In the random partition, clients were assigned a mixture vector generated randomly based on $Uniform(0, 1)$. The EMNIST dataset was divided into $K$ clusters, ensuring no data overlap between them.
\textbf{Synthetic CIFAR10:} This synthetic dataset mirrored CIFAR10's partitions but had different training and testing data distributions to simulate dynamic client data. In the 10:90 partition, 50 clients had $90\%$ training data with $10\%$ testing data from $D_{A}$ and vice versa. The rationale for separate training and testing data distributions is explained in Appendix~\ref{subsubsec:synthetic_dataset_insights}.

\subsection{Focus of Experimental Study}
\label{subsec:Focus_of_Experimental_Study}

First, we compare the clustering ability of {\ipfl} with a recent clustering-based pFL algorithm~\cite{Ruan2022FedSoftSC}. Second, we compare {\ipfl} with other non-clustering pFL models with a simple test accuracy evaluation. Taking it one step further, we provide a comparison of {\ipfl} and other clustering and non-clustering pFL models in terms of opt-out reduction and PMA maintenance. Lastly, we show that including client preferences while clustering yields better personalization results because clients can make decisions based on knowledge restricted to the aggregator server. We evaluate all algorithms using the datasets that were employed in their original publications. Thus ensuring that we highlight the strengths of each algorithm, rather than skewing the evaluation towards our methodology.

\begin{table*}[t]
\vspace{-0.15cm}
\begin{minipage}{0.49\textwidth}
\centering
\caption{Clusters models accuracy (CIFAR10)}
\vspace{-0.23cm}
\begin{adjustbox}{width=1\linewidth}

\begin{tabular}{lcccccccc}
\toprule
\addlinespace[-0.5pt] 
\vspace{-5pt}
& \multicolumn{4}{c}{10:90} & \multicolumn{4}{c}{30:70} \\
\vspace{-5pt}
& \multicolumn{2}{c}{$\theta_{0}$} & \multicolumn{2}{c}{$\theta_{1}$} & \multicolumn{2}{c}{$\theta_{0}$} & \multicolumn{2}{c}{$\theta_{1}$} \\
\vspace{-1.9pt}
& c0 & c1 & c0 & c1 & c0 & c1 & c0 & c1 \\
\midrule
\addlinespace[-0.1pt] 
\vspace{-2pt}
\textbf{IP-FL} & \textbf{63.7} & 41.3 & 43.7 & \textbf{63.8} & 58 & 57.7 & \textbf{58.6} & \textbf{58.8} \\
\vspace{-2pt}
\textbf{FedSoft} & 48.9 & 49.5 & \textbf{50.7} & \textbf{49.6} & 48 & 48.4 & \textbf{50} & \textbf{50} \\\vspace{-1pt}
\textbf{IFCA} & 46.4 & 44.3 & \textbf{47} & \textbf{45} & 46 & 44 & \textbf{47.4} & \textbf{46.5} \\
\vspace{-2pt}
\textbf{FedEM} & 45.7 & 46.5 & \textbf{46.8} & \textbf{47.1} & 43.8 & 45.2 & \textbf{46.3} & \textbf{47.4} \\
\addlinespace[-1pt] 
\bottomrule
\vspace{-2pt}
\end{tabular}

\end{adjustbox}
\label{tab:CIFAR10_tier_model_acc_clustering_algos}
\end{minipage}%
\hspace{2pt}
\begin{minipage}{0.5\textwidth}
\vspace{-0.6em}
\caption{Test accuracy of pFL methods (EMNIST)}
\vspace{-0.6em}
\fontsize{20}{10}\selectfont 
\centering
\begin{adjustbox}{width=1\linewidth}
\begin{tabular}{@{}lcccccc@{}}
\toprule
\addlinespace[5pt] 
Partitions & Ditto & FedProx & FedALA & PerfFedAvg & FedProto & {\ipfl} \\ 
\addlinespace[5pt] 
\midrule
\addlinespace[5pt] 
\addlinespace[5pt] 
10:90 & 85.8±4.8 & 75.2±4.8 & 75.5±4.7 & 87.5±3.8 & 72±1.4 & \textbf{87.5±3.7} \\
\addlinespace[5pt] 
30:70 & 76±4.5 & 79.7±4 & 78.4±3.2 & 76.6±3.9 & 59.7±4.7 & \textbf{85.1±3.4} \\
\addlinespace[5pt] 
Linear & 75.3±5.1 & 82.8±2.7 & 82±3.6 & 80.8±3.5 & 62.6±4.9 & \textbf{83.4±4.9} \\
\addlinespace[5pt] 
Random & 77.8±6.8 & 80.9±4.4 & 79±5.1 & 83.3±5.2 & 68.4±5.7 & \textbf{86.2±4.3} \\ 
\addlinespace[5pt] 
\bottomrule
\end{tabular}
\end{adjustbox}
\label{tab:pFL_test_accuracy_comparisons}
\end{minipage}
\vspace{-1.5em}
\end{table*}

\begin{wrapfigure}{r}{0.5\textwidth}
\vspace{-2.5em}
\centering
    \begin{subfigure}{0.49\linewidth}
        \centering
        \includegraphics[width=\linewidth]{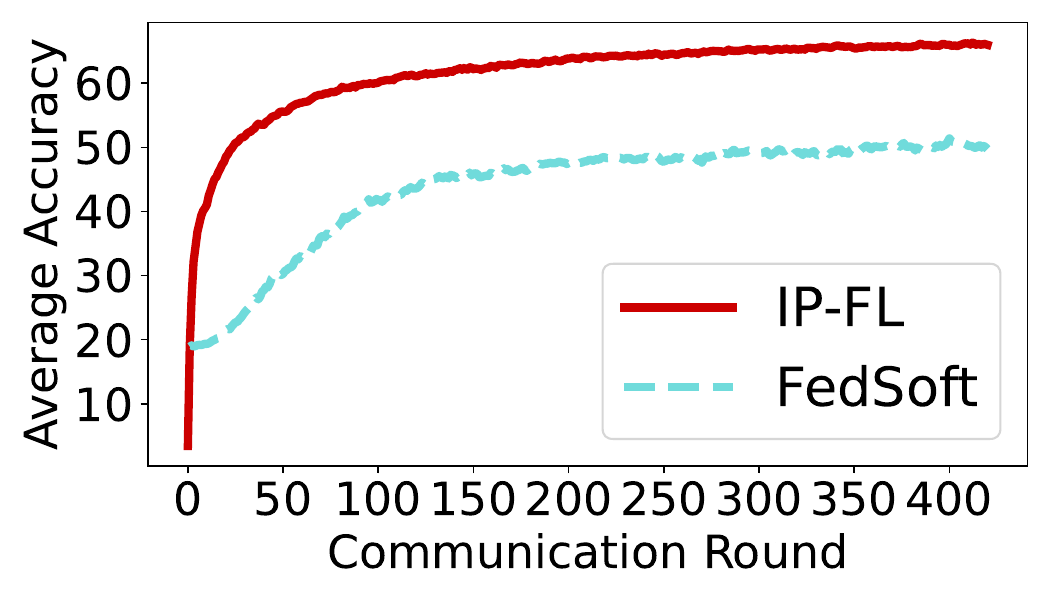}
        \label{fig:FedSoft_10_90_convergence_speedup}
    \end{subfigure}
    \begin{subfigure}{0.49\linewidth}
        \centering
        \includegraphics[width=\linewidth]{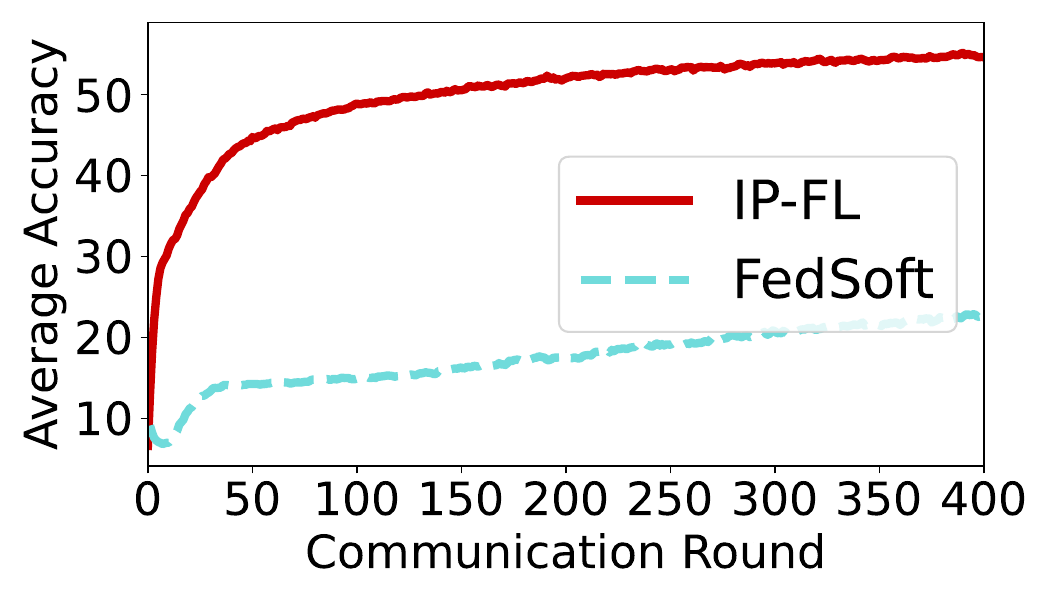}
        \label{fig:FedSoft_30_70_convergence_speedup}
    \end{subfigure}
    \includegraphics[width=\linewidth]{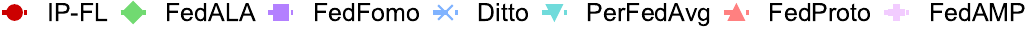}
    \label{acc_legend}
    \begin{subfigure}{0.485\linewidth}
        \centering
        \includegraphics[width=\linewidth]{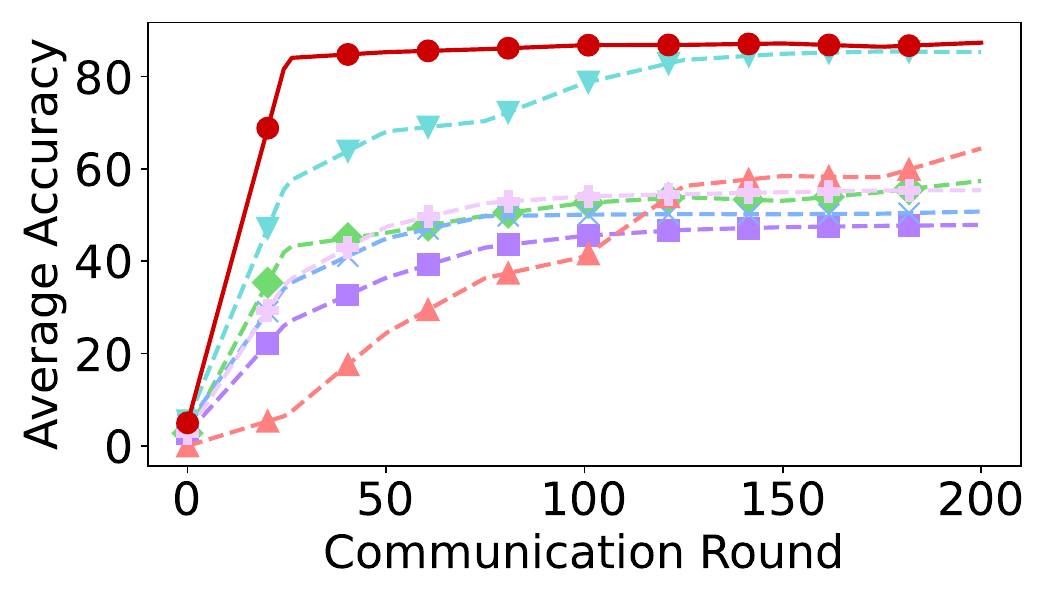}
        \caption{10:90 partition}
        \label{subfig:all_pFL_10_90_convergence_speedup}
    \end{subfigure}
    \begin{subfigure}{0.485\linewidth}
        \centering
        \includegraphics[width=\linewidth]{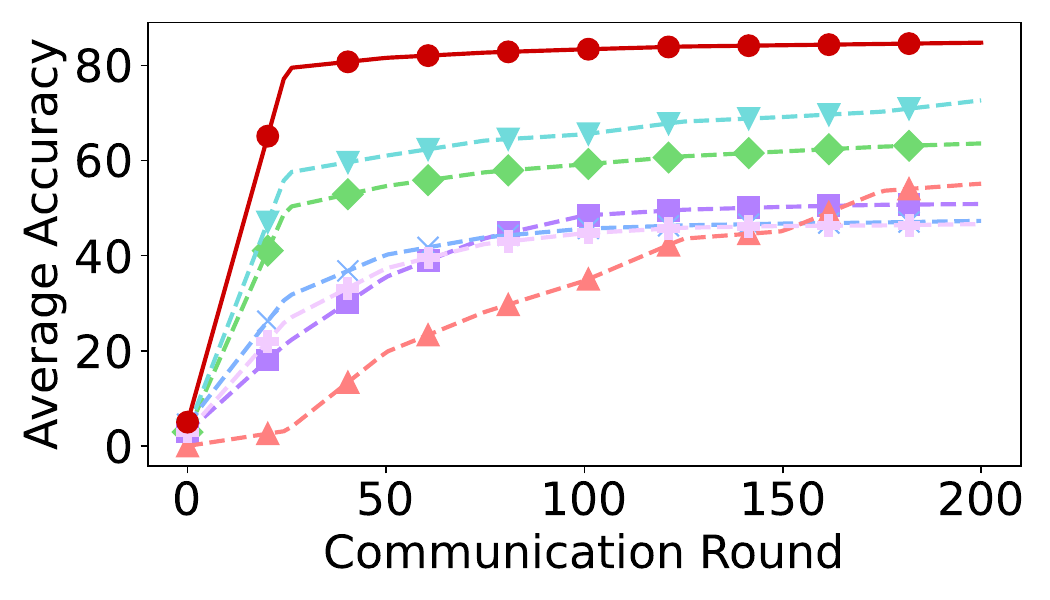}
        \caption{30:70 partition}
        \label{subfig:all_pFL_30_70_convergence_speedup}
    \end{subfigure}

    \caption{Convergence Speed: {\ipfl} vs. other pFL algorithms}
    \vspace{-1em}
    \label{fig:all_pFL_convergence_speed}
\end{wrapfigure}

\subsection{\bf Test Accuracy Performance Study}
\label{subsec:Results_Synthetic_data}
\textbf{Effectiveness of clustering.}
We evaluate cluster-level model performance on holdout datasets from cluster distributions ($D_{A}$ and $D_{B}$), comparing {\ipfl} with recent cluster-based pFL algorithm FedSoft on CIFAR10, matching experiment settings from~\cite{Ruan2022FedSoftSC} with 100 clients, batch size 128, learning rate $\eta=0.01$, and 300 training rounds.
Table \ref{tab:synthetic_PI_FL_and_FedSoft} presents test accuracy for two data partitions: \textbf{10:90} and \textbf{30:70}, using {\ipfl}. Notably, {\ipfl} outperforms FedSoft in the 10:90 partition, where each cluster represents a singular distribution. Clients with more $\theta_{0}$ data train in cluster $c_{0}$ with $63.68\%$ accuracy, and clients with more $\theta_{1}$ data prefer cluster $c_{1}$ with $63.82\%$ accuracy.
In contrast, FedSoft scores lower accuracies of $50.7\%$ and $49.6\%$ in the 10:90 partition, facing challenges with data partition variability and representing singular distributions. FedSoft's models $c_{0}$ and $c_{1}$ show similar performance across $\theta_{0}$ and $\theta_{1}$ data, indicating its personalization is more effective when the majority of clients have share data, hence struggling with individual distribution representation and underperforming in non-IID scenarios. The 30:70 partition shows diminished performance, attributed to less data heterogeneity, which constrains personalization-driven accuracy gains.
In Tables~\ref{tab:Synthetic_CIFAR10_tier_model_acc_clustering_algos} and~\ref{tab:CIFAR10_tier_model_acc_clustering_algos}, we compare {\ipfl} with three clustering-based pFL algorithms: FedSoft, IFCA, and FedEM~\cite{marfoq2021federated_multitask}. {\ipfl} consistently outperforms all three algorithms, supporting its effectiveness.

\textbf{Comparison with non-clustering pFL models.} Table~\ref{tab:pFL_test_accuracy_comparisons} shows the test accuracy comparison of {\ipfl} with other recent pFL algorithms. Some pFL models can perform well for individual partitions such as Ditto for 10:90, FedProx and FedALA for Linear, and PerFedAvg for Random, however, {\ipfl} consistently outperforms for all data partitions.


\textbf{Convergence speed. }
Figure~\ref{fig:all_pFL_convergence_speed} shows {\ipfl}'s superior convergence speed and personalized accuracies over other pFL methods, matching the communication rounds per training epoch of clustering-based algorithms~\cite{IFCA,Ruan2022FedSoftSC}. This efficiency stems from incorporating client preferences into post-aggregation updates, avoiding the additional proximal update step needed in clustering-based approaches. Moreover, {\ipfl} enables optional single-shot personalization with cluster models during training or post-training, further enhancing efficiency.

\begin{figure*}[t]
\centering
{\includegraphics[width=1.0\textwidth]{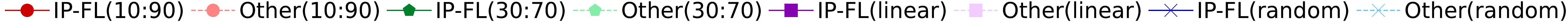} \label{FedSoft_data_dist_legend}}
    \begin{subfigure}[htbp]{0.27\linewidth}
        \centering
        \includegraphics[width=1.0\linewidth]{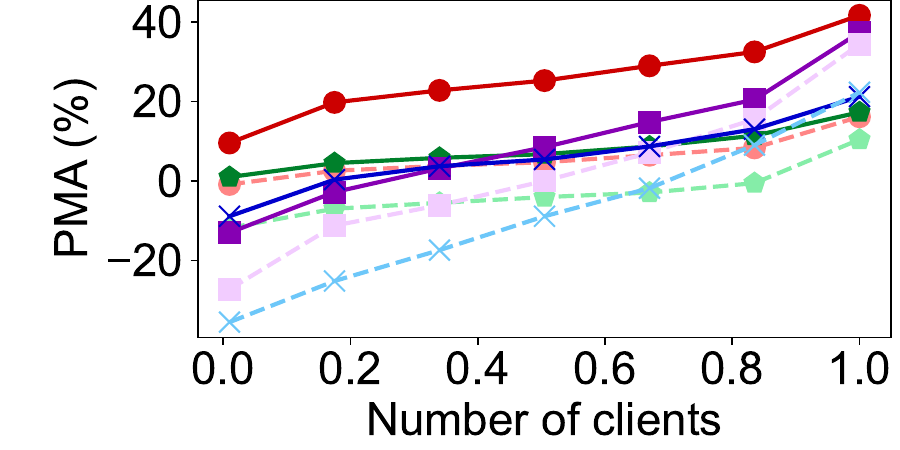}
        \caption{{\ipfl} \& FedSoft}
        \label{fig:FedSoft_FedAvg_PIFL_Cifar10_personalized_acc}
    \end{subfigure}
    \begin{subfigure}[htbp]{0.23\linewidth}
        \includegraphics[width=1.0\linewidth]{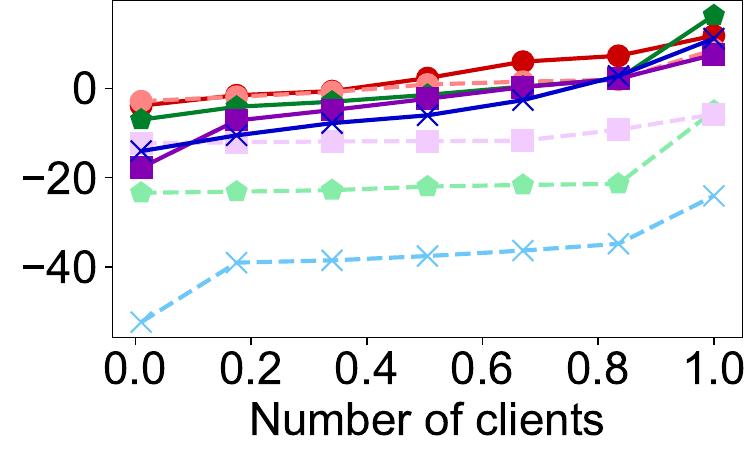}
        \caption{{\ipfl} \& Ditto}
        \label{fig:Ditto_FedAvg_PIFL_personalized_acc}
    \end{subfigure}
     \begin{subfigure}[htbp]{0.23\linewidth}
        \includegraphics[width=1.0\linewidth]{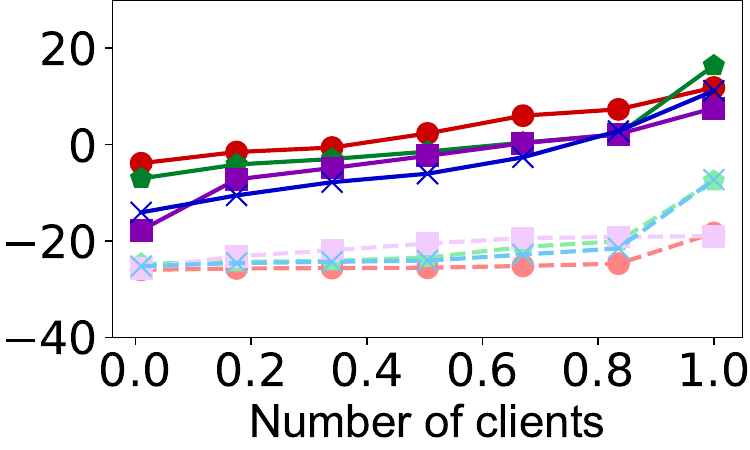}
        \caption{{\ipfl} \& FedALA}
        \label{fig:FedALA_FedAvg_PIFL_personalized_acc}
    \end{subfigure}
    \begin{subfigure}[htbp]{0.23\linewidth}
        \includegraphics[width=1.0\linewidth]{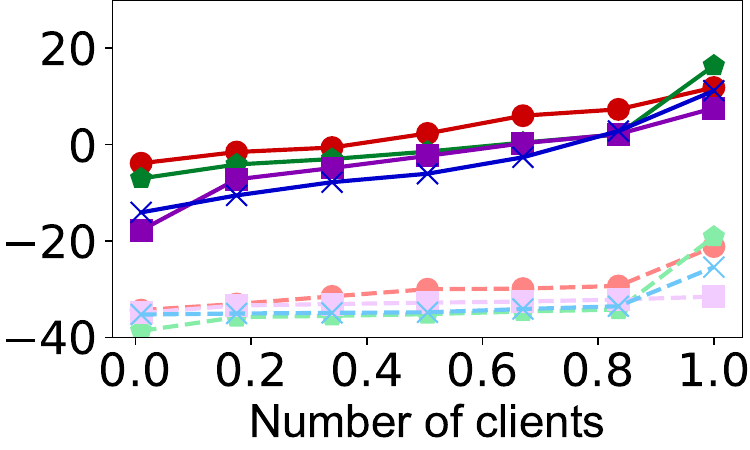}
        \caption{{\ipfl} \& FedFomo}
        \label{fig:FedFomo_FedAvg_PIFL_personalized_acc}
    \end{subfigure}
    \begin{subfigure}[htbp]{0.28\linewidth}
        \includegraphics[width=1.0\linewidth]{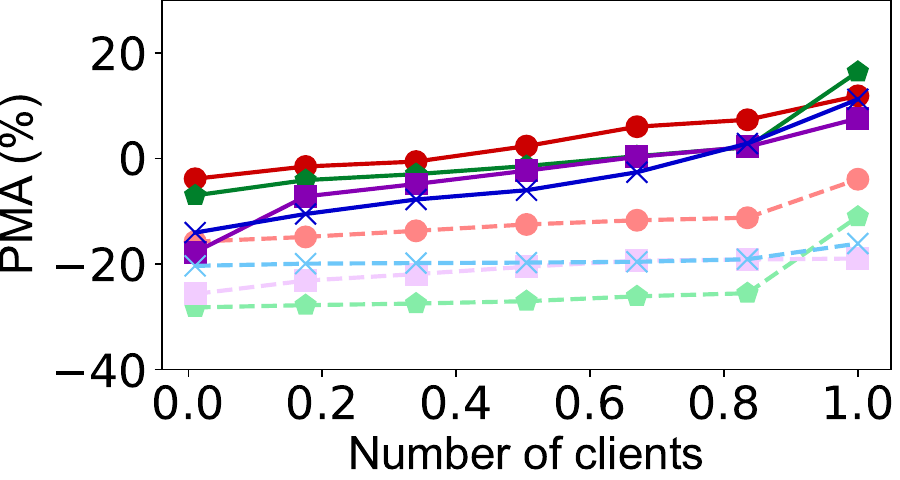}
        \caption{{\ipfl} \& FedProto}
        \label{fig:FedProto_FedAvg_PIFL_personalized_acc}
    \end{subfigure}
    \begin{subfigure}[htbp]{0.25\linewidth}
        \includegraphics[width=1.0\linewidth]{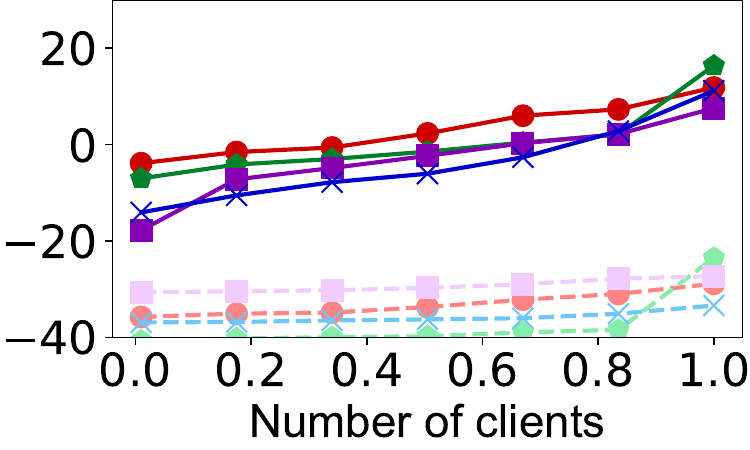}
        \caption{{\ipfl} \& FedAMP}
        \label{fig:FedAMP_FedAvg_PIFL_personalized_acc}
    \end{subfigure}
    \begin{subfigure}[htbp]{0.12\linewidth}
        \includegraphics[width=1.0\linewidth]{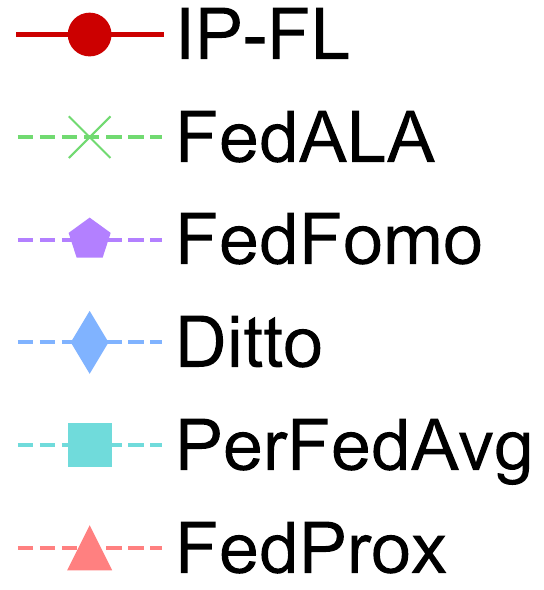}
        \label{fig:classes_legend}
    \end{subfigure}
    \begin{subfigure}[htbp]{0.28\linewidth}
        \includegraphics[width=1.0\linewidth]{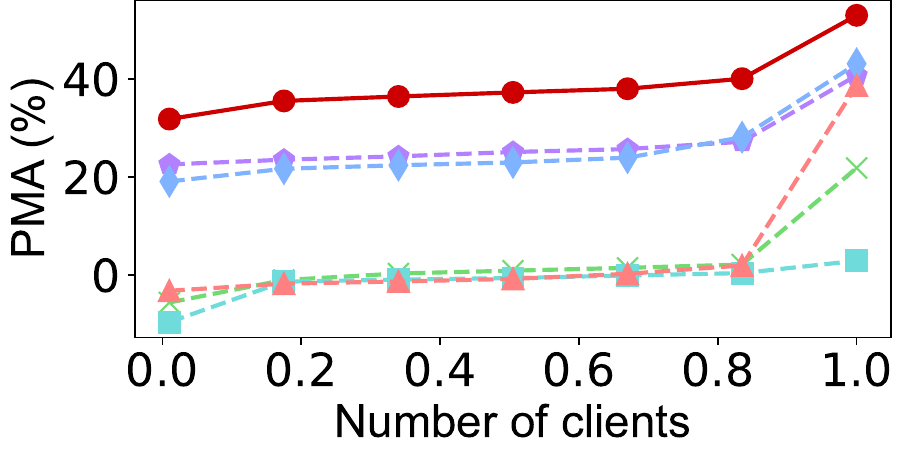}
        \caption{4 classes per client}
        \label{fig:PIFL_all_pFL_4_classes_personalized_acc}
    \end{subfigure}
    \vspace{-4pt}
    \caption{CDF of clients’ PMA for different datasets and methods}
    \vspace{-2em}
\label{fig:personalized_model_appeal}
\end{figure*}

\vspace{-1.1em}
\begin{wraptable}{r}{0.7\textwidth}
\vspace{-1.7em}
\centering
\caption{PMA and opt-outs with the feature-shift EMNIST}
\vspace{-5pt}
\label{tab:pFL_PMA_optout_accuracy_comparisons_feature_shift}
\resizebox{\linewidth}{!}{
  \begin{tabular}{lcccccccc}
  \toprule
  \textbf{pFL algo} & \textbf{FedAvg} & \textbf{Ditto} & \textbf{FedProx} & \textbf{FedALA} & \textbf{FedFomo} & \textbf{FedProto} & \textbf{PerfFedAvg} & \textbf{{\ipfl}} \\ 
  \midrule
  \textbf{Pers. Acc.} & 75.7±3 & 83.7±2 & 78.8±14 & 83.3±3 & 63.9±10 & 46.5±21 & 83.7±3 & \textbf{84.3±2} \\
  \textbf{Optouts} & - & 0 & 23 & 0 & 100 & 0 & 0 & \textbf{0} \\
  \textbf{Avg. PMA} & - & 7.6 & 3 & 7.2 & -17 & -48.6 & 7.7 & \textbf{8.6} \\ 
  \bottomrule
  \end{tabular}
}
\end{wraptable}
\subsection{\bf Experimental Study of Opt-outs and PMA}
\label{subsec:effectiveness_in_optouts_and_PMA}

Figure~\ref{fig:personalized_model_appeal} shows the CDF of PMA for clients using different datasets and pFL methods. {\ipfl} excels, especially for the 10:90 partition where data heterogeneity is most prominent (Figure~\ref{fig:FedSoft_FedAvg_PIFL_Cifar10_personalized_acc}). In contrast, the EMNIST dataset which features a broader class distribution per client allows FedAvg to perform relatively well, suggesting limited scope for further improvement via personalization.
{\ipfl} enhances PMA, particularly in the 10:90 and 30:70 partitions where other pFL solutions struggle. 
Table~\ref{tab:pFL_PMA_optout_accuracy_comparisons} in Appendix shows a highly heterogeneous scenario with 52 clusters and 4 classes per client, Ditto and FedProto perform well, however {\ipfl} outperforms by $15\%$ in PMA. Opt-out ratios for FedProx, FedALA, and PerFedAvg are $0.64$, $0.31$, and $0.68$, respectively, with no opt-outs for Ditto, FedFomo, and {\ipfl}, showcasing {\ipfl}'s effectiveness in reducing opt-outs and improving PMA under high data heterogeneities.


\begin{wrapfigure}{r}
{0.3\linewidth}
\vspace{-2em}
  \centering
  \includegraphics[width=1\linewidth]{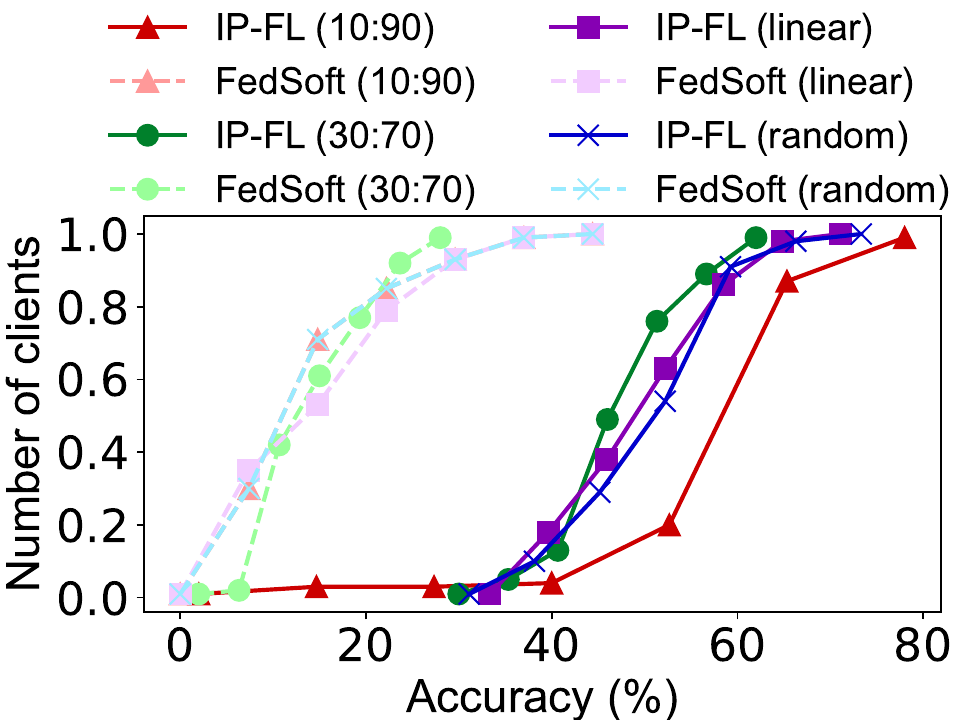}
  \vspace{-5pt}
  \caption{{\ipfl} and FedSoft with Synthetic CIFAR10 data}
  \vspace{-10pt}
\label{fig:FedSoft_PIFL_CIFAR10_personalized_acc}
\end{wrapfigure}
\textbf{PMA and opt-outs with Feature-shift.}
We compared {\ipfl} to other pFL algorithms in non-IID data scenarios using a feature shift technique, with identical hyperparameters. Two datasets were created: $D_{A}$ with normal EMNIST images and $D_{B}$ with images rotated by 90 degrees. Results in Table~\ref{tab:pFL_PMA_optout_accuracy_comparisons_feature_shift} show that {\ipfl} that {\ipfl} surpasses other pFL models, improving test accuracy by 1--38\% and PMA by 12-118\% with feature-shifted non-IID data.

\textbf{Advantages of including client preferences in pFL.}
{\ipfl} maintains personalized model test accuracy, even with dynamic client data or the inadvertent addition of a new client to the wrong cluster. In Figure~\ref{fig:FedSoft_PIFL_CIFAR10_personalized_acc}, we observe the CDF of clients' test accuracy after 500 training rounds. {\ipfl}'s robustness to varying client data is evident, while FedSoft, relying on the server's perspective without access to client data, struggles to make precise clustering decisions.

\begin{wrapfigure}{r}
{0.3\linewidth}
\vspace{-4em}
  \centering
  \includegraphics[width=1\linewidth]{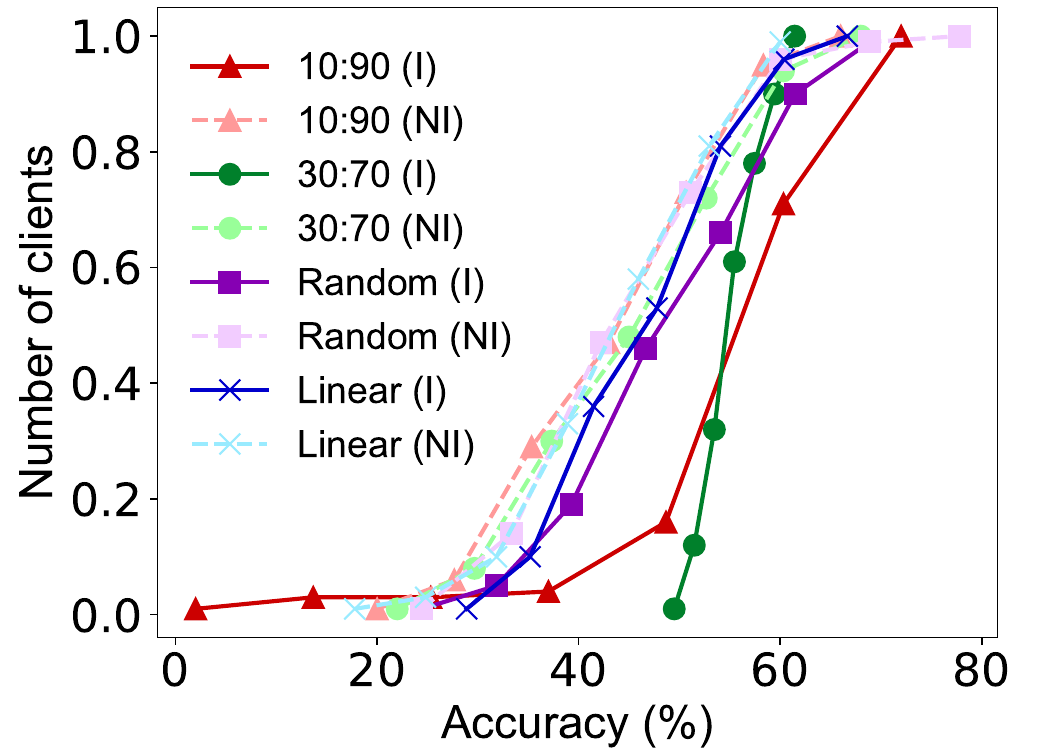}
  \vspace{-5pt}
  \caption{{\ipfl} with and without incentive (I/NI)}
  \vspace{-20pt}
  \label{fig:PI-FL_ablatian_study}
\end{wrapfigure}

\textbf{Ablatian study with Incentive in {\ipfl}. }
In Figure~\ref{fig:PI-FL_ablatian_study}, we compared {\ipfl} with (I) and without (NI) incentives, and the results show a higher accuracy with incentives enabled under a majority of data scenarios suggesting the importance of incentives in pFL.

\vspace{-6pt}
\section{Conclusion}
\label{sec:Conclusion}
 \vspace{-6pt}
In this paper, we proposed {\ipfl} to address the challenges of incentive provision in pFL for increasing consistent participation by providing appealing personalized models to clients. {\ipfl} client-centric clustering approach ensures accurate clustering and improved performance even in case of dynamic data distribution shift of the client's local data or inadvertently mistaken clustering decision by the client. Unlike prior works that consider incentivizing and personalization as separate problems, {\ipfl} solves them as interrelated challenges yielding improvement in pFL performance. Extensive empirical evaluation shows its promising performance compared to other state-of-the-art works.

\bibliography{main}
\bibliographystyle{plainnat}

\appendix

\section{{\ipfl} Usecases}
\label{sec:PI-FL Usecases}

\subsection{Personalized and Incentives Learning Applications}

Incentivizing participants in personalized federated learning can guide their behavior towards joining appropriate or similar subgroups, enhancing the personalization and effectiveness of models. Here are a few use cases demonstrating this:

\textbf{Educational Content Personalization:} {\ipfl} can be used to gather learning behavior data from students. Incentives encourage students to share data, helping to cluster them based on learning styles and progress, leading to more tailored educational content and methods.

\textbf{Energy Consumption Optimization in Smart Homes:} Homeowners can be incentivized to share energy usage data. This data helps create models that predict and optimize energy consumption patterns in different household clusters, promoting energy efficiency~\cite{smart_cities_2}.

\textbf{Precision Agriculture:} Farmers can be motivated to share crop and soil data. Clustering this data helps develop models that provide personalized farming advice, enhancing crop yields and sustainable practices~\cite{SINGH2022380_agriculture}.

\textbf{Healthcare Data from Wearables:} Incentives encourage users of wearable health devices to share specific health data. This data can then be clustered into subgroups like heart patients or diabetes management, leading to more personalized healthcare recommendations~\cite{smart_healthcare}.

\textbf{Driving Data in Autonomous Vehicles:} Vehicle owners can be incentivized to share their driving data. This data can be used to cluster drivers into groups based on driving styles or habits, which can then be used to improve autonomous driving systems tailored to different driving behaviors~\cite{federated_vehicle_transportation,XIAO2021107338_human_behavior}.

\textbf{Mobile Device Usage:} Users can be incentivized to share data for services like app recommendations. By clustering users with similar app usage patterns, more personalized and relevant app suggestions can be made~\cite{mobileNetworks}.

In each scenario, {\ipfl} not only boosts participation rates but also ensures that contributions are more aligned with specific subgroup needs, leading to highly personalized and effective solutions.

\subsection{Responsible Personalized learning over BigData through incentives}
As per the recent PRIVACY-PRESERVING DATA
SHARING AND ANALYTICS (PPDSA) initiative from the US government~\cite{WhiteHouse2023Privacy} the aim is to balance the need for leveraging big data for societal benefits (like healthcare, security, or economic growth) with the imperative of protecting individual privacy rights. Personalized Federated Learning enables the collaborative use of data in a way that inherently respects privacy concerns and can be used for personalized objective learning of individual groups in large populations. However, the collection of this data for personalized learning is only possible with the proper incentivization of each population group.

\subsection{AI Marketplaces and role of {\ipfl}}
\label{subsec:AI Marketplaces and role of PI-FL}

The AI field is witnessing the emergence of several AI marketplaces like ModelPlace~\cite{modelplace}, GravityAI~\cite{gravityai}, AWS AI Marketplace~\cite{awsaimarketplace}, and AI Marketplace~\cite{aimarketplace}, serving both model and data requirements. A crucial distinction, however, is while these platforms offer data and model training services, their data is often static and not as contemporary as data generated on individual client devices. There are also pronounced challenges related to data transportation, such as significant costs and privacy issues.

For applications rooted in user recommendations and driven by user data, assimilating the latest user trends is of paramount significance. Personalized models through federated learning offer an effective strategy to realize this goal. However, the current federated model marketplace is limited, primarily due to the lack of incentives encouraging clients or users to share their recent and valuable data, which can be pivotal in developing top-tier models.

Our work is meticulously curated to bridge this gap, championing the design of architectures and frameworks that stimulate this sector's growth. By accentuating the importance of contemporary and pertinent data from client devices, our objective is to create a framework that incentivizes active user engagement, leading to the production of higher-caliber models. This initiative is grounded in the belief that motivating clients/users to share their data is essential to harness federated learning's complete potential in dynamic, personalized applications.

\section{Initial Clustering}
\label{sec:Initial Clustering}

In addition to the analyses presented in the "Experimental Study" section, we also include an evaluation of the 40:60 partition using the EMNIST dataset, following the hyperparameters outlined in the "Experimental Study" section. Notably, in this new partition setting, our proposed {\ipfl} algorithm performs better compared to other algorithms under consideration. As mentioned in the prior response 40:60 is slightly heterogeneous compared to the complete IID case of 50:50 which is why we see an improved performance with personalization particularly with {\ipfl}.

 {\ipfl} also has a mode that facilitates clients in forming initial clusters. This helps clients bypass early decisions, optimizing spending when contributions and cluster distributions are ambiguous and similarity metrics that the clients use among other metrics to make clustering decisions are unknown. Initially, the clients perform pre-training for a few rounds, through evaluation we have found that by pre-training for just 5 rounds the clients can be profiled. After this, the profiler calculates per-class F1-Scores $\xi$ on an IID test dataset \cite{scikit-f1scores}. The next training round's client clustering uses the K-Means algorithm \cite{scikit-Kmeans} based on the most varied F1-scores $V_{F1}$ from $C$ classes, as per Equation \ref{eq:imprtant_features}.

\begin{equation}
\label{eq:imprtant_features}
V_{F1} = var(\xi_{i}) \in [1, C] \mid \forall i \in N
\end{equation}

Our evaluations exclude this feature, but {\ipfl} offers it for rapid convergence and cost-saving. Realizing the constraints in choosing all clients for pre-training, only clients replying within a threshold are used for F1 score calculations. Others are assigned to clusters randomly, later settling into suitable clusters through preference and contribution.

\section{Fairness in {\ipfl}}
\textbf{Consumer and Provider Roles:} The incentive mechanism in {\ipfl} treats clients as both providers and consumers. As a consumer, the client tries to attain a certain level of personalized model appeal, so it pays the provider to spend resources to participate in training for the said model in each round. Whereas as a provider, the client earns a profit based on its marginal contribution to training the cluster models.

\textbf{Fairness for consumers:} The clients with consumer’s profile pay for the tokens they use. To ensure that the consumers only pay for what they are getting they have the option of opt-out, where consumer clients can opt-out if they are not getting a model with a PMA over the desired threshold. To overcome the coarse-grained limitation of opt-out, the consumer clients also get a partial reimbursement of their tokens spent using the reimbursement if the resulting PMA is not up to their expected threshold.

\textbf{Opportunity fairness:} We have followed other state of art token-based incentive federated learning frameworks that ensure no consumer or a sub-group of consumers is able to manipulate the overall training process by spending more tokens which is why we have a single pricing model for all consumers. Starting with the same number of tokens also ensures equal opportunity for all consumer clients.

\textbf{Fairness for providers:} We also ensure collaboration fairness among clients with the provider’s profile. A provider that has more marginal contribution (calculated using Shapley Values) in training the model receives more token rewards as highlighted in the Algorithms presented in the paper. The second incentive for providers similar to consumers is increased PMA. Both rewards are dependent on accurate clustering choices from individual provider clients. Hence, to achieve these incentives each client makes better clustering choices which lead to clusters that have similar client objectives (good personalization) and this increases the provider client’s incentives (PMA and tokens). This is theoretically supported by the theoretical analysis presented in the paper.







\section{Further evaluation of the empirical performance of {\ipfl}}

\begin{table*}[h]
\centering
\caption{Test accuracy of ablation study with Incentive (I) and without incentive (NI)}
\resizebox{0.8\textwidth}{!}{%
\label{tab:PI-FL_ablatian_study}
\begin{tabular}{@{}lcccccccc@{}}
\toprule
& \multicolumn{2}{c}{\textbf{10:90}} & \multicolumn{2}{c}{\textbf{30:70}} & \multicolumn{2}{c}{\textbf{linear}} & \multicolumn{2}{c}{\textbf{random}} \\
& \textbf{c0} & \textbf{c1} & \textbf{c0} & \textbf{c1} & \textbf{c0} & \textbf{c1} & \textbf{c0} & \textbf{c1} \\ \midrule
{\ipfl} (I) & \multicolumn{1}{c}{$58.62 (0)$} & \multicolumn{1}{c}{$67.4 (1)$} & \multicolumn{1}{c}{$51.12 (0)$} & \multicolumn{1}{c}{$64.06 (1)$} & \multicolumn{1}{c}{$66.2 (0)$} & \multicolumn{1}{c}{$57.02 (1)$} & \multicolumn{1}{c}{$64.54 (0)$} & \multicolumn{1}{c}{$56.86 (1)$} \\
{\ipfl} (NI) & \multicolumn{1}{c}{$49.92 (1)$} & \multicolumn{1}{c}{$52.8 (1)$} & \multicolumn{1}{c}{$48.9 (0)$} & \multicolumn{1}{c}{$44.44 (1)$} & \multicolumn{1}{c}{$50.82 (1)$} & \multicolumn{1}{c}{$45.92 (1)$} & \multicolumn{1}{c}{$57.42 (1)$} & \multicolumn{1}{c}{$46.76 (1)$} \\ \bottomrule
\end{tabular}
}
\end{table*}

\subsection{\textbf{Ablatian study of clustering performance with Incentive in {\ipfl}}}
\label{subsec:ablatian_study_clustering}

We perform an ablation study with the incentive component of {\ipfl} on the Synthetic CIFAR10 dataset for 200 rounds with $N=100$ clients, batch size 128, and learning rate $\eta=0.01$. When the incentive is disabled clients do not consider maximizing their incentive while sending preference bids. Instead, clients send preference bids with random cluster choices to the scheduler as in FedAvg~\cite{McMahan_FL}.

Table \ref{tab:PI-FL_ablatian_study} shows the test accuracies of the cluster-level models. {\ipfl}(I) indicates that incentives are enabled and PI-F(NI) shows the accuracies when incentives are disabled. In general {\ipfl}(I) outperforms {\ipfl}(NI) in terms of test accuracy for all partitions. The important point to note here is that the incentive mechanism in {\ipfl} directly motivates clients to join clusters in which they can make the most contribution. This results in accurate clustering based on client data distributions and good-quality personalized models. This is indicated by the performance of {\ipfl}(NI), i.e without incentive, cluster-level models are unable to dominate a single distribution and only perform well for a single distribution for all partitions except 30:70. Compared to this, in {\ipfl}(I) each cluster-level model dominates and performs well for their distribution.


\begin{figure}
  \begin{center}
    \includegraphics[width=0.4\textwidth]{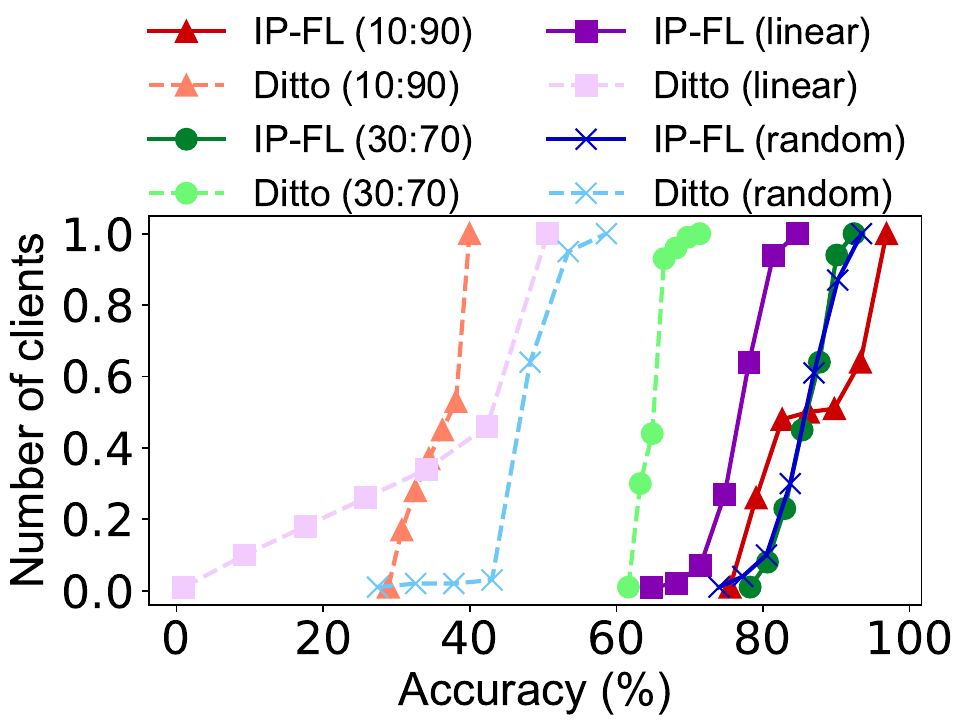}
  \end{center}
  \caption{CDF of clients' personalized model test accuracy for {\ipfl} and Ditto on EMNIST Dataset}
  \label{fig:PI-FL_Ditto}
\end{figure}

\subsubsection{Multiple distribution Results}
\label{Multiple distribution Results}

\subsection{\bf Advantages of including client preferences in pFL.}
\label{subsec:including_client_preferences}

\subsubsection{Challenges requiring client autonomy in pFL}
\label{Challenges requiring client autonomy}

Prior personalized FL works generate personalized models from the server's perspective. We argue that the server may not have complete information to produce good-quality models due to a variety of challenges \cite{kairouz2019advances,li2020federated}. These challenges are as follows: \textbf{Confidentiality}: Some clients may have sensitive data that they do not want to share with others for privacy or security reasons. For example, a company may have confidential customer data that they do not want to share with a third-party vendor, \textbf{Competitive Advantage}: In some industries, companies may want to keep their data private to maintain a competitive advantage. For example, a company may not want to share its sales data with competitors. \textbf{Data Governance}: Some organizations may have strict data governance policies that prohibit the sharing of certain types of data. For example, a healthcare organization may not be able to share patient data without proper consent. \textbf{Resource limitations}: Clients with large datasets may not have the resources to share all their data for training. In these cases, they may choose to share a random sample of their data to keep the training process manageable. \textbf{Data Anonymization}: Sometimes, clients may not want to share the raw data, but instead, they may share a subset of the data which has been anonymized to protect the privacy of the individuals.
\textbf{Compliance with privacy laws}: To comply with privacy laws (GDPR \cite{regulation2018general}, HIPA \cite{act1996health}) some clients might only share anonymized data while keeping Personally Identifiable Information (PII) private. Prior clustering-based pFL works also lack support to accurately include new clients into the clusters whose data qualities are unknown. By including client preferences, {\ipfl} performs accurate clustering and generation of appealing personalized models for new clients.

We use the Synthetic datasets to test {\ipfl} where the aggregator is unaware of the client's dataset distribution and goals.

\begin{table*}[h]
\centering
\caption{Test accuracy for {\ipfl} on Synthetic CIFAR10 Dataset}
\label{tab:PI-FL_CIFAR10}
\resizebox{0.6\textwidth}{!}{%
\begin{tabular}{@{}lcccccccc@{}}
\toprule
& \multicolumn{2}{c}{\textbf{10:90}} & \multicolumn{2}{c}{\textbf{30:70}} & \multicolumn{2}{c}{\textbf{linear}} & \multicolumn{2}{c}{\textbf{random}} \\
& \textbf{c0} & \textbf{c1} & \textbf{c0} & \textbf{c1} & \textbf{c0} & \textbf{c1} & \textbf{c0} & \textbf{c1} \\ \midrule
$\theta_{0}$ & \multicolumn{1}{c}{\textbf{62.78}} & \multicolumn{1}{c}{2.56} & \multicolumn{1}{c}{\textbf{53.28}} & \multicolumn{1}{c}{34.32} & \multicolumn{1}{c}{\textbf{61.66}} & \multicolumn{1}{c}{12.08} & \multicolumn{1}{c}{\textbf{66.9}} & \multicolumn{1}{c}{19.48} \\
$\theta_{1}$ & \multicolumn{1}{c}{1.5} & \multicolumn{1}{c}{\textbf{70.96}} & \multicolumn{1}{c}{30.38} & \multicolumn{1}{c}{\textbf{61.94}} & \multicolumn{1}{c}{30.94} & \multicolumn{1}{c}{\textbf{59.44}} & \multicolumn{1}{c}{19.12} & \multicolumn{1}{c}{\textbf{58.42}} \\ \bottomrule
\end{tabular}%
}
\end{table*}

\begin{table*}[th]
\centering
\caption{Test accuracy for {\ipfl} and FedSoft on Synthetic CIFAR10 Dataset}
\label{tab:PI-FL_FedSoft_CIFAR10}
\resizebox{0.8\textwidth}{!}{%
\begin{tabular}{@{}lcccccccc@{}}
\toprule
& \multicolumn{2}{c}{\textbf{10:90}} & \multicolumn{2}{c}{\textbf{30:70}} & \multicolumn{2}{c}{\textbf{linear}} & \multicolumn{2}{c}{\textbf{random}} \\
& \textbf{c0} & \textbf{c1} & \textbf{c0} & \textbf{c1} & \textbf{c0} & \textbf{c1} & \textbf{c0} & \textbf{c1} \\ \midrule
{\ipfl} & \multicolumn{1}{c}{$62.68 (0)$} & \multicolumn{1}{c}{$70.96 (1)$} & \multicolumn{1}{c}{$53.28 (0)$} & \multicolumn{1}{c}{$61.94 (1)$} & \multicolumn{1}{c}{$61.66 (0)$} & \multicolumn{1}{c}{$59.44 (1)$} & \multicolumn{1}{c}{$66.90 (0)$} & \multicolumn{1}{c}{$58.42 (1)$} \\
FedSoft & \multicolumn{1}{c}{$32.50 (0)$} & \multicolumn{1}{c}{$38.62 (1)$} & \multicolumn{1}{c}{$20.28 (0)$} & \multicolumn{1}{c}{$23.58 (0)$} & \multicolumn{1}{c}{$34.42 (1)$} & \multicolumn{1}{c}{$49.62 (1)$} & \multicolumn{1}{c}{$21.62 (1)$} & \multicolumn{1}{c}{$33.12 (1)$} \\ \bottomrule
\end{tabular}%
}
\end{table*}

\begin{table*}[h]
\centering
\caption{Test accuracy for FedSoft on Synthetic CIFAR10 Dataset}
\label{tab:FedSoft_CIFAR10}
\begin{tabular}{@{}lcccccccc@{}}
\toprule
& \multicolumn{2}{c}{\textbf{10:90}} & \multicolumn{2}{c}{\textbf{30:70}} & \multicolumn{2}{c}{\textbf{linear}} & \multicolumn{2}{c}{\textbf{random}} \\
& \textbf{c0} & \textbf{c1} & \textbf{c0} & \textbf{c1} & \textbf{c0} & \textbf{c1} & \textbf{c0} & \textbf{c1} \\ \midrule
$\theta_{0}$ & \multicolumn{1}{c}{\textbf{32.5}} & \multicolumn{1}{c}{13.6} & \multicolumn{1}{c}{\textbf{20.28}} & \multicolumn{1}{c}{\textbf{23.58}} & \multicolumn{1}{c}{8.48} & \multicolumn{1}{c}{2.82} & \multicolumn{1}{c}{16.18} & \multicolumn{1}{c}{0.28} \\
$\theta_{1}$ & \multicolumn{1}{c}{11.76} & \multicolumn{1}{c}{\textbf{38.62}} & \multicolumn{1}{c}{0.18} & \multicolumn{1}{c}{0.08} & \multicolumn{1}{c}{\textbf{34.42}} & \multicolumn{1}{c}{\textbf{49.62}} & \multicolumn{1}{c}{\textbf{21.62}} & \multicolumn{1}{c}{\textbf{33.12}} \\ \bottomrule
\end{tabular}
\end{table*}

\subsubsection{Clustering performance with Synthetic CIFAR10 dataset}
\label{clustering_performance_Synthetic_CIFAR10}

Cluster-based pFL methods cluster all heterogeneous clients within different clusters so each cluster has homogeneous clients with similar data distribution in it. So first we test the client-preference driven clustering design of {\ipfl} with Synthetic CIFAR10 Data to show the performance of cluster-level models. We use the same configurations described in the main paper for the CIFAR10 dataset evaluation. We perform training for 500 rounds with both {\ipfl} and FedSoft. Table \ref{tab:PI-FL_CIFAR10} shows the cluster-level model test accuracies for {\ipfl} with Synthetic CIFAR10 data.

{\ipfl} accurately differentiates between clients of different distributions.
This is visible by the accuracy difference of each cluster-level model on different distributions. For example on the 10:90 partition $c1$ model has a $70.96\%$ accuracy on the $\theta_{1}$ distribution and has $2.56\%$ accuracy on $\theta_{0}$ which indicates that cluster-level model $c1$ trains with clients that have the majority of their training data from $\theta_{1}$. Similarly, $c0$ trains with clients that have their majority of training data from $\theta_{0}$ and has an accuracy of $62.68\%$.

Table \ref{tab:PI-FL_FedSoft_CIFAR10} shows the cluster-level model accuracy comparison of {\ipfl} and FedSoft. 
The number inside the parenthesis along with accuracy shows the distribution for which the cluster-level model performs best. For example, for the linear partition, {\ipfl} $c0$ cluster-level model has an accuracy of $61.66\%$ for distribution $\theta_{0}$ and $59.44\%$ accuracy for distribution $\theta_{1}$. For the linear partition with FedSoft, both $c0$ and $c1$ perform best on only one distribution $\theta_{1}$ with accuracy $34.42\%$ and $49.62\%$. 
{\ipfl} outperforms FedSoft in terms of accuracy for each partition and can distinguish between different distributions accurately.

Table \ref{tab:FedSoft_CIFAR10} represents the result of FedSoft \cite{Ruan2022FedSoftSC} tested with the Synthetic CIFAR10 dataset. The experimental setup is explained in more detail in the main paper. We can observe that except for the 10:90 partition, FedSoft is unable to cluster clients under individual clusters in a way that each cluster-level model could dominate one distribution. Both cluster-level models perform well on only one distribution and the other is ignored leading to low test accuracy of both cluster-level models for the ignored distribution. 




\begin{figure}[htb]
\centering
\centerline{\includegraphics[width=0.4\textwidth]{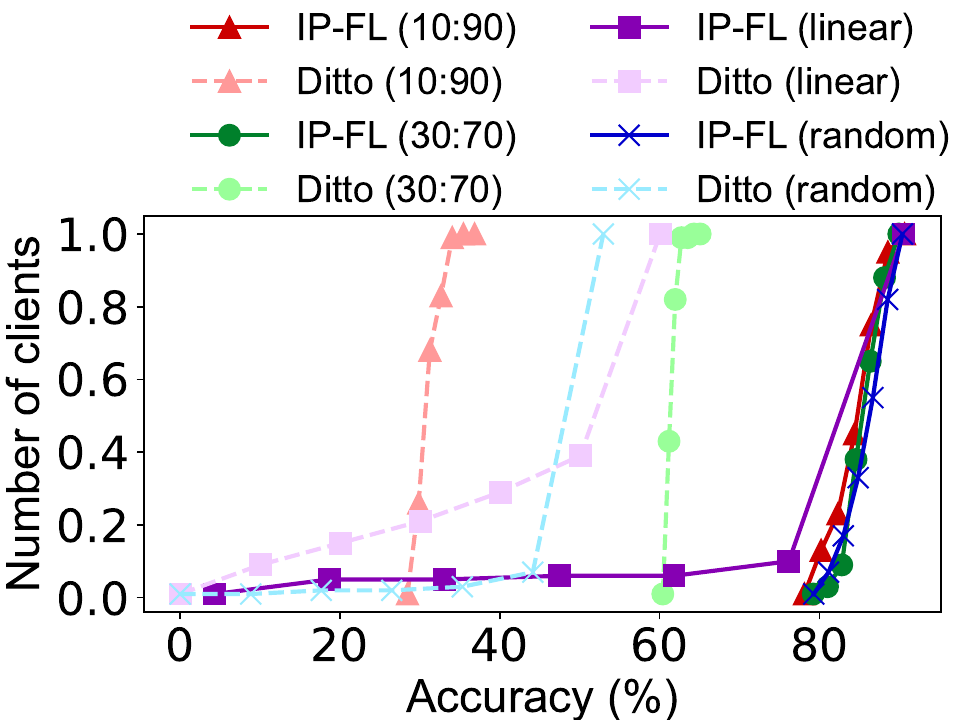}}
\caption{CDF of clients' personalized model test accuracy for {\ipfl} and Ditto on 4 and more distributions with EMNIST Dataset}
\label{fig:PI-FL_Ditto_4_distributions}

\end{figure}

\subsubsection{Clustering performance with Synthetic EMNIST dataset}
\label{clustering_performance_Synthetic_EMNIST}

 This image dataset has images of dimension 28 x 28 and 52 output classes where 26 classes are lower case letters and 26 classes are upper case letters. We test the dataset on different partitions of 10:90, 30:70, linear, and random created in the same way as the Synthetic CIFAR10 data. The only difference is that $D_{A}$ contains 26 lowercase letters and $D_{B}$ has 26 uppercase letters.

Since Ditto is not a clustering-based algorithm, we only compare the test accuracy of personalized models on the EMNIST dataset for {\ipfl} and Ditto in Figure \ref{fig:PI-FL_Ditto}. {\ipfl} outperforms Ditto significantly, especially for highly heterogeneous data partitions such as 10:90 and linear. The reason for this performance improvement is that, unlike Ditto, {\ipfl} provides autonomy to clients to personalize according to their own goals. With Ditto, the goal of each client which consists of their private data is hidden from the aggregator server which affects the quality of personalized models.

{\ipfl} outperforms other FL personalization algorithms in heterogeneous cases when the clients and dataset are divided between 2 distributions (DA \& DB). To analyze the reliability of {\ipfl} it is also evaluated in highly heterogeneous conditions with 4 different distributions (DA, DB, DC, and DD) on the EMNIST dataset. EMNIST dataset has a total of 56 classes, thus each of the 4 distributions gets $25\%$ of total available classes. DA has the first 13 classes, DB has the next 13, and so on. Figure \ref{fig:PI-FL_Ditto_4_distributions} shows the CDF of test accuracy for all personalized models at clients. {\ipfl} outperforms Ditto for all partition types. For the linear partition less than $10\%$ clients have lower than $80\%$ accuracy and we attribute this as an outlier due to the partition type where dividing the data linearly some clients get very few data samples. This trend can also be seen with Ditto for the linear partition.




\subsubsection{Insights from the analysis with Synthetic datasets}
\label{subsubsec:synthetic_dataset_insights}

We use the synthetic datasets to present a scenario of dynamic data at client or induction of new clients in pFL whose data distributions are unknown. This presents new challenges of accurate clustering and the generation of personalized models for less familiar clients. We observe through empirical evaluation with Synthetic datasets that this can lead to $100\%$ opt-outs from these clients if we use conventional server-driven pFL methods. However, we observe that by the induction of client preferences in clustering, {\ipfl} can do accurate clustering and generate appealing models for new clients and changing data distribution at clients.

\subsection{Opt-out results for 4 classes per client with EMNIST dataset}

Table~\ref{tab:pFL_PMA_optout_accuracy_comparisons} shows the complete opt-out, test accuracy, and PMA results for the test conducted with the EMNIST dataset with 4 classes per client. Some pFL algorithms such as FedFomo and Ditto have similar performance compared to {\ipfl} in terms of opt-outs. However, {\ipfl} outperforms all of them in terms of PMA and test accuracy.

\begin{figure}[htb]
  \begin{center}
    \includegraphics[width=0.4\textwidth]{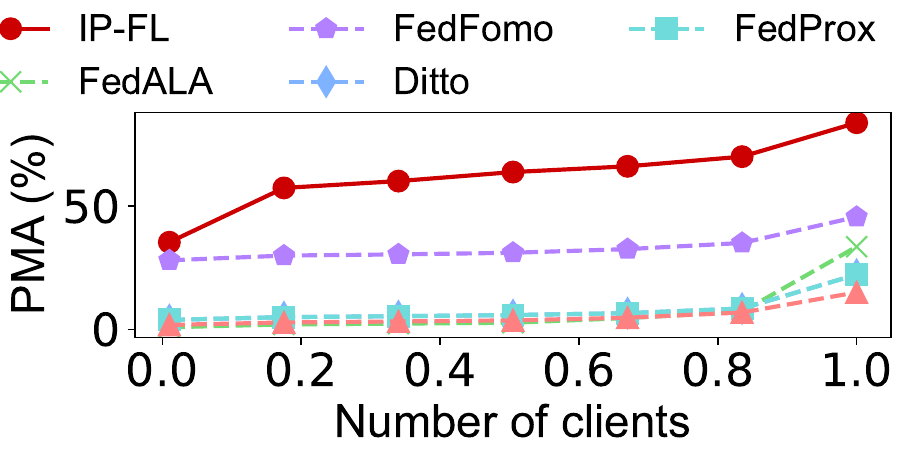}
  \end{center}
  \caption{CDF of clients' PMA for {\ipfl} and pFL algorithms on CIFAR10 Dataset with 2 classes per client}
  \label{fig:PI-FL_pFL_2_classes}
\end{figure}

\subsection{Opt-out results for 2 classes per client with CIFAR10 dataset}

We also test highly heterogeneous data conditions where each client has only 2 classes per client. We use the learning rate = 0.01, batch size = 128, and global epochs = 150 with the CIFAR10 dataset and the same CNN model mentioned in the main paper. Figure~\ref{fig:PI-FL_pFL_2_classes} shows the PMA for different pFL algorithms. {\ipfl} outperforms all other pFL algorithms by approximately $50\%$ and FedFomo by $30\%$. This goes to show that {\ipfl} performs even better in highly heterogeneous conditions where personalization is used to tackle data heterogeneity.

\begin{table*}[htb]
\centering
\caption{PMA and opt-outs with the 4 classes per client EMNIST dataset}
\label{tab:pFL_PMA_optout_accuracy_comparisons}
\resizebox{0.9\textwidth}{!}{%
\begin{tabular}{@{}lccccccc@{}}
\toprule
pFL Method & FedAvg & Ditto & FedProx & FedALA & PerFedAvg & FedFomo & {\ipfl} \\ \midrule
Personalized Accuracy & 82.8±4.2 & 90.78±2.1 & 61.23±2.86 & 62.17±3.9 & 59.99±1.87 & 86.57±1.59 & \textbf{98.59±1.2} \\
Optouts & - & 0 & 0.64 & 0.31 & 0.68 & 0 & \textbf{0} \\
Average PMA & - & 24.48±4.2 & 0.57±5.14 & 1.5±4.74 & -0.6±1.5 & 25.9±3.6 & \textbf{37.93±3.43} \\ \bottomrule
\end{tabular}
}
\end{table*}

\begin{table*}[hb]
\centering
\caption{PMA and opt-outs with 40:60 partition of the EMNIST dataset}
\label{tab:40:60_EMNIST}
\begin{adjustbox}{width=0.9\textwidth}
\begin{tabular}{@{}lccccccc@{}}
\toprule
pFL Method & FedAvg & Ditto & FedProx & FedALA & PerFedAvg & FedFomo & {\ipfl} \\
\midrule
Personalized Accuracy & $76.22 \pm 0.47$ & $77.93 \pm 1.1$ & $82.2 \pm 0.47$ & $74.65 \pm 3.51$ & $79.14 \pm 2.36$ & $47.65 \pm 14$ & \textbf{85.11 ± 2.1} \\
Optouts & - & 0 & 26 & 0 & 100 & 0 & \textbf{0} \\
Average PMA & - & 1.7 & 5.9 & 0.14 & 2.9 & -28.6 & \textbf{8.86} \\
\bottomrule
\end{tabular}
\end{adjustbox}
\end{table*}

\subsection{Opt-out results for slightly non-IID EMNIST dataset}
We also evaluate the 40:60 partition using the EMNIST dataset, following the parameters outlined in the section~\ref{subsec:effectiveness_in_optouts_and_PMA}. The results for this are shown in Table~\ref{tab:40:60_EMNIST}. Notably, in this new partition setting, our proposed {\ipfl} algorithm performs better compared to other algorithms under consideration. The 40:60 partition is slightly heterogeneous compared to the complete IID case of 50:50 which is why we see an improved performance with personalization particularly with {\ipfl}.

\begin{figure*}[htb]
\centering
    {\includegraphics[width=0.6\textwidth]{PI-FL/Figures/legend_EMNIST_all-cropped.pdf} \label{acc_legend} }
    \vfil
\begin{subfigure}[h]{0.3\textwidth}
    \centering
\includegraphics[width=1\linewidth]{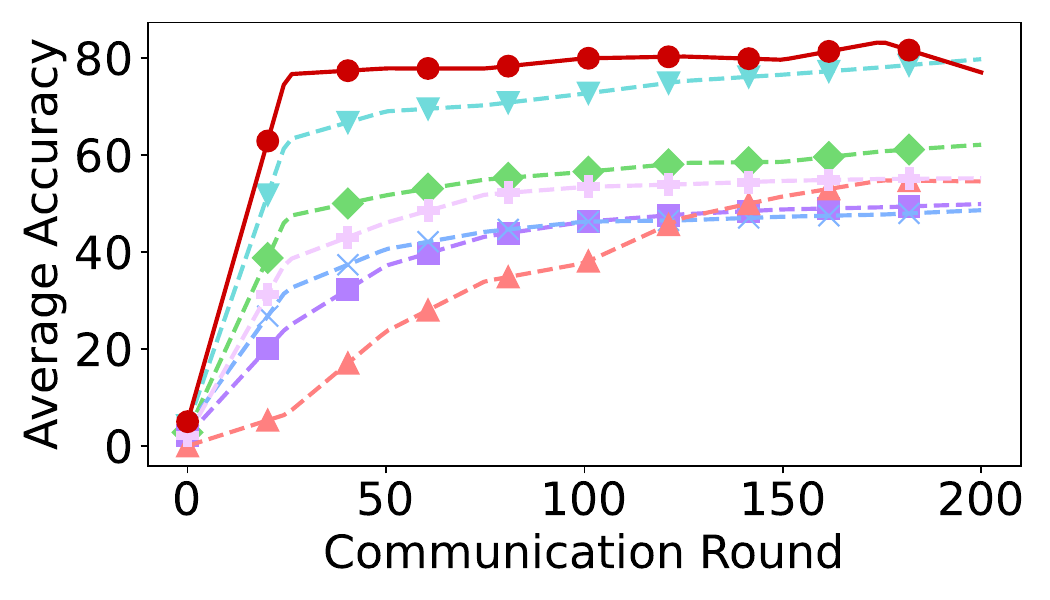}
\caption{Linear partition}
    \label{subfig:all_pFL_linear_convergence_speedup}
\end{subfigure}
\begin{subfigure}[h]{0.3\textwidth}
    \centering
\includegraphics[width=1\linewidth]{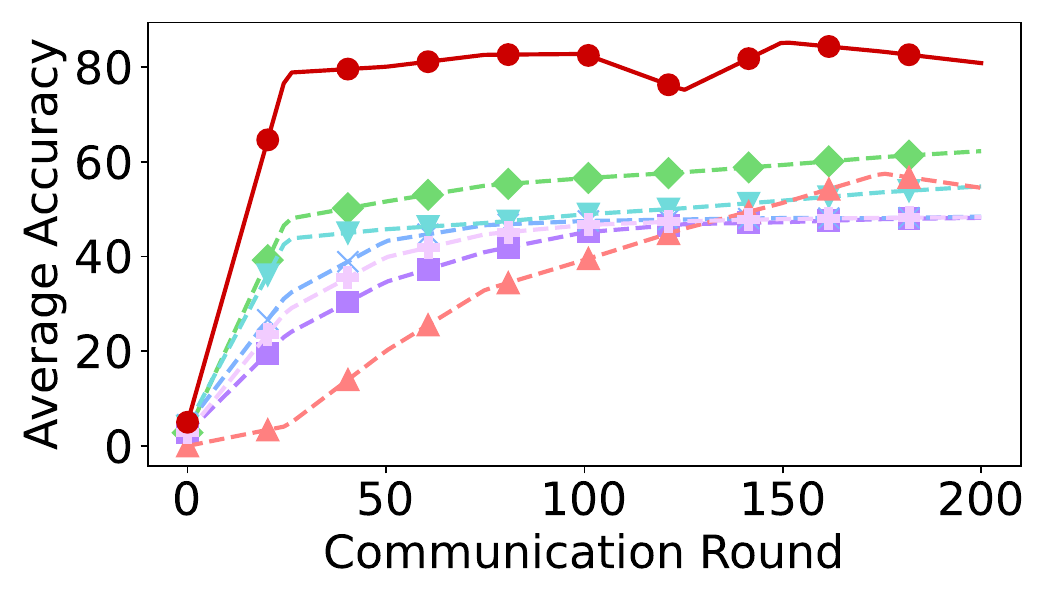}
\caption{Random partition}
    \label{subfig:all_pFL_random_convergence_speedup}
\end{subfigure}
\caption{Convergence Speed of other pFL algorithms vs. {\ipfl}}
\label{fig:all_pFL_convergence_speed_appendix}
\end{figure*}

\section{Convergence Speed}
\label{Convergence Speed}

In this section, we include the remaining results of convergence speed from the experimental evaluation. Figure~\ref{fig:all_pFL_convergence_speed_appendix} shows the convergence speed of {\ipfl} in comparison with other recent pFL algorithms. The results clearly indicate that {\ipfl} outperforms all other pFL algorithms in terms of both accuracy achieved as well as time to convergence for all datasets.

\section{Cost and Overhead}
In terms of computation costs, {\ipfl} vastly outperforms clustering-based pFL algorithms. The significant advantage arises because other recent cluster-based pFL solutions~\cite{Ruan2022FedSoftSC} necessitate an additional update step during training, referred to as proximal update. On the other hand, {\ipfl} clients only require a single-shot personalization using cluster models, which is not only efficient but also conveniently performed by each client once the training concludes. It is worth noting that this personalization step is not mandatory during training, which further adds to the efficiency of our approach. 

\section{Limitations}

Calculating Shapley values for contribution calculation puts some additional overhead on the aggregator server however, we have partially resolved that by using Shapley Values approximations. In the future, we can easily include more lightweight contribution calculation mechanisms as they evolve. In addition to that, calculating evaluation metrics such as PMA could consume resources, however, we would emphasize that PMA is an evaluation metric employed to demonstrate performance improvements of {\ipfl} over other pFL algorithms. PMA or threshold calculation is not required for the {\ipfl} algorithm itself. We have just use it to compare with other pFL algorithms. Therefore, in the practical application of {\ipfl}, calculating PMA or the threshold is not a requisite step. Even if we take local training as the threshold the results will be the same because the threshold only serves the purpose of comparing the {\ipfl} algorithm with other pFL algorithms.

\section{Shapley value approximation for client contribution}
\label{ShapleyValueApprox}

Here we present the Shapley Value approximation derivation we use for calculating the client contributions.

  	\begin{algorithm}[tb]
		\centering
		\caption{Estimated Shapley value of any client in an FL}
            \label{alg:ShapleyValues}
		\footnotesize
		\begin{algorithmic}[1]
			\renewcommand{\algorithmicrequire}{\textbf{Input:}} 
			\renewcommand{\algorithmicensure}{\textbf{Output:}}
			
			\REQUIRE 
			Test data $(x_i,y_i),i=1,\ldots,\ntest$, clients' local model parameters and aggregation weights, $\param_m,\lambda_m$. server's aggregated model parameter $\param_M = \sum_{m=1}^{M} \lambda_m \param_m$.
			
			\STATE 
			Calculate $\gamma_M \de  \ntest^{-1} \sum_{i=1}^{\ntest} \nabla_{\param} \ell(x_i, y_i; \param_{M}) $
			
			\FOR{$k=1, \ldots, M$}
			\STATE Calculate $\shap(i \rightarrow [M] )$ using 
			\begin{gather}
			    - \biggl( \frac{1}{\ntest} \sum_{i=1}^{\ntest} \nabla_{\param} \ell(x_i, y_i; \param_{M}) \biggr)^\T \weight_i \param_i
       \end{gather}
       \textrm{ for unnormalized aggregation, or}
       \begin{gather}
			    - \biggl( \frac{1}{\ntest} \sum_{i=1}^{\ntest} \nabla_{\param} \ell(x_i, y_i; \param_{M}) \biggr)^\T \weight_i (\param_i -\param_M)
			\end{gather}
                \textrm{ for normalized aggregation}.
			\ENDFOR
			\ENSURE Obtains all clients' Shapley values
		\end{algorithmic}
	\end{algorithm}
 
\textbf{Notation}: Let $[M]$ denote the set $\{1,\ldots,M\}$, and $A-B$ the set of elements in $A$ but not in $B$.
In this section, $[M]$ denotes all the agents that \textit{participate} in the coalition. We will consider those not participating in the near future.

We aim to look for a reasonable way to quantify the amount of each client's contribution in a round. 
Suppose at any particular round, the server obtains an aggregated model with parameter
\begin{align}
    \param_{[M]} \de \sum_{m \in [M]} \weight_m \param_m , \label{eq_thetaM}
\end{align}
where $\weight_m$ is the weight (usually $n_m/n$ where $n_m$ and $n$ are sample sizes of client $m$ and all clients, respectively), and $\param_m$ is the locally updated model of client $m$.

The prediction loss of the model with parameter $\param$, denoted by $\loss(\param)$, is approximated by
\begin{align}
    \loss(\param) \approx \frac{1}{\ntest} \sum_{i=1}^{\ntest} \ell(x_i, y_i; \param),
\end{align}
where $(x_i, y_i)$, $i=1,\ldots,\ntest$, is a set of test data.
At round $t$, we define the value function of a set of agents $C$ based on how much their contributed model, denoted by $\param_C$, has decreased the loss of the earlier model, denoted by $\param_{t-1}$, namely  
\begin{align}
    \Value_t(C) \de  \loss(\param_{t-1})-\loss(\param_C), \label{eq_value2}
\end{align}
so that the larger the better.
When there is no ambiguity, we simply write $\Value_t$ as $\Value$.
It is worth noting that $\Value$ is a function of the set while $\loss$ is a function of the parameter. Once $C$ is realized, $\param_C$ will become $\param_t$ for the next round.

Recall that the original Shapley value~\cite{roth1988introduction} of agent $m$ given a set of agents $A$ and a value function $v$ is defined by 
\begin{align}
    &\sum_{S \in A-\{m\}} \frac{|S|! (|A|-1-|S|)!}{|A|!} ( v(\{S \cup \{m\}\}) - v(S) ),  \label{eq_Shapley}
\end{align}
whose sum over all agents is equal to $v(A) - v(\emptyset)$. Here, $\emptyset$ represents the \textit{baseline} coalition scenario, from which the contribution of each agent is quantified.
To highlight the dependency on baseline, we use $B$ to denote the baseline and rewrite (\ref{eq_Shapley}) as 
\begin{equation}
\begin{aligned}
    \shap(m \rightarrow A \mid B) 
    &= \sum_{S \in A-\{m\}} \frac{|S|! (|A|-1-|S|)!}{|A|!} \times \\
    &\quad ( v(\{S \cup \{m\}\} \mid B) - v(S \mid B) ),
\end{aligned}
\label{eq_Shapley2}
\end{equation}
where $v( S \mid B)$ means the value of $S$ conditional on the baseline $B$. In our scenario, $B$ means the set of agents that are already in coalition and thus 
\begin{align}
    v( S \mid B) \de v(S \cup B). \label{eq_value}
\end{align}
Let us consider the baseline as $B \de [M]-\{i,j\}$. The corresponding baseline model will be

\begin{equation}
\begin{aligned}
   &\textrm{unnormalized version:} \\ 
   &\param_{[M]-\{i,j\}} \de \sum_{m \in [M]-\{i,j\}} \weight_m \param_m , \label{eq_thetaij}
\end{aligned}
\end{equation}

\begin{equation}
\begin{aligned}
   &\textrm{normalized version:} \\
   &\param_{[M]-\{i,j\}}^* \de \frac{1}{\sum_{m \in [M]-\{i,j\}} \weight_m}  \sum_{m \in [M]-\{i,j\}} \weight_m \param_m . \label{eq_6}
\end{aligned}
\end{equation}
We consider an unnormalized version for brevity. 
The additional value by introducing $i,j$ is 
\begin{align}
    &\Value(\{i,j\} \mid [M]-\{i,j\}) - \Value(\emptyset \mid [M]-\{i,j\}) \\
    &\substack{\textrm{use} (\ref{eq_value})}=\Value([M]) - \Value([M]-\{i,j\}) \\
    &\substack{\textrm{use} (\ref{eq_value2}) \textrm{ and recall } (\ref{eq_thetaM}) and (\ref{eq_thetaij})}= 
    \loss(\param_{[M]-\{i,j\}}) - \loss(\param_M)\label{eq_4}  \\
    &=\frac{1}{\ntest} \sum_{i=1}^{\ntest} \{\ell(x_i, y_i; \param_{M}) - \ell(x_i, y_i; \param_{[M]-\{i,j\}})\} \nonumber \\ 
    &\approx \frac{1}{\ntest} \sum_{i=1}^{\ntest} \nabla_{\param} \ell(x_i, y_i; \param_{M})^\T (\param_{[M]-\{i,j\}}-\param_{M}) \label{eq_7}\\
    &= - \biggl( \frac{1}{\ntest} \sum_{i=1}^{\ntest} \nabla_{\param} \ell(x_i, y_i; \param_{M}) \biggr)^\T (\weight_i \param_i + \weight_j \param_j) . \label{eq_3}
\end{align}
Next, we calculate how much agent $i$ should be attributed to the above gain that is achieved by $i,j$ jointly. 
To that end, we calculate the Shapley value of agent $i$ conditional on that agents in $[M]-\{i,j\}$ already participate, namely
\begin{align}
    &\shap(i \rightarrow \{i,j\} \mid  [M]-\{i,j\}) \\
    &\substack{\textrm{recall }(\ref{eq_Shapley2})}=\sum_{S \in \{j\}} \frac{|S|! (1-|S|)!}{2!} \biggl( v\biggl(S \cup \{i\} \cup ([M]-\{i,j\})\biggr)   - v\biggl(S \cup ([M]-\{i,j\}) \biggr) \biggr)\nonumber \\
    &~~~~~~~~~~~~~= \frac{1}{2} \biggl( v([M]) - v([M]-\{j\}) + v([M]-\{i\}) - v([M]-\{i,j\}) \biggr) \nonumber \\
    &~~~~~~~~~~~~~= \frac{1}{2} \biggl( -\loss(\param_{M}) + \loss(\param_{M, -i}) - \loss(\param_{M, -j}) + \loss(\param_{M, -ij}) \biggr) \nonumber
    \end{align}
    \begin{align}
    &= \frac{1}{2} \biggl( -\loss(\param_{M}) + \loss(\param_{M, -i}) + \loss(\param_{M})  - \loss(\param_{M, -j}) + \loss(\param_{M, -ij}) - \loss(\param_{M}) \biggr)\label{eq7} \\
    &\substack{\textrm{use } (\ref{eq_4})-(\ref{eq_3}) \textrm{ and alike }}\approx \\ 
    &-\frac{1}{2}\biggl( \frac{1}{\ntest} \sum_{i=1}^{\ntest} \nabla_{\param} \ell(x_i, y_i; \param_{M}) \biggr)^\T \nonumber \\
    &\hspace{1cm}\cdot (\weight_i \param_i - \weight_j \param_j + \weight_i \param_i + \weight_j \param_j) \nonumber \\
    &= - \biggl( \frac{1}{\ntest} \sum_{i=1}^{\ntest} \nabla_{\param} \ell(x_i, y_i; \param_{M}) \biggr)^\T \weight_i \param_i ,\label{eq_Svalue}
\end{align}
which, interestingly, does not depend on $j$.
As such, we use this to calculate the Shapley value of client~$i$, denoted by
\begin{align}
    &\shap(i \rightarrow [M] ) 
    \de  - \biggl( \frac{1}{\ntest} \sum_{i=1}^{\ntest} \nabla_{\param} \ell(x_i, y_i; \param_{M}) \biggr)^\T \weight_i \param_i. \label{eq_Svalue2}
\end{align}

From Equalities~(\ref{eq_3}) and~(\ref{eq_Svalue}), we can verify that \begin{align}
    &\Value(\{i,j\} \mid [M]-\{i,j\}) - \Value(\emptyset \mid [M]-\{i,j\})\\
    &=\shap(i \rightarrow [M] )  + \shap(j \rightarrow [M] ) \nonumber
\end{align}   

\begin{remark}[Intuitions]\label{remark_shap1}
    Intuitively, our derived Shapley value of client $i$ in  (\ref{eq_Svalue}) can be regarded as the model's marginal reduction of the test loss by introducing client $i$. To see that, consider the following approximation based on first-order Taylor expansion:
    \begin{align}
        &\frac{1}{\ntest} \sum_{i=1}^{\ntest} \ell(x_i, y_i; \param_{M}-\Delta)  - \frac{1}{\ntest} \sum_{i=1}^{\ntest} \ell(x_i, y_i; \param_{M}\param) \nonumber \\
        &\approx
        - \biggl(\frac{1}{\ntest} \sum_{i=1}^{\ntest} \nabla_{\param} \ell(x_i, y_i; \param_{M}) \biggr)^\T \Delta\param, \nonumber
    \end{align}
    which becomes the term in (\ref{eq_Svalue}) when $\Delta\param \de \weight_i \param_i$.
    The above quantity approximates the amount of client $i$'s contribution to decreasing the test loss of the server's aggregated model, the larger the better.
\end{remark}

\begin{remark}[Normalized counterpart] \label{remark_shap2}
Suppose we use the normalized version introduced in (\ref{eq_6}) when considering the baseline without clients $i,j$. Thus,
\begin{align}
    \param^*_{[M]-\{i,j\}}
    &=
    \frac{\param_{M} - \lambda_i \param_i - \lambda_j \param_j}{\sum_{m \in [M]-\{i,j\}} \weight_m}  \nonumber \\
    &=\param_{M} + \frac{(\lambda_i+\lambda_j)\param_{M} - \lambda_i \param_i - \lambda_j \param_j }{\sum_{m \in [M]-\{i,j\}} \weight_m} \nonumber\\
    &=\param_{M} - \frac{\lambda_i (\param_{i} - \param_{M}) + \lambda_j (\param_{j} - \param_{M}) }{1-(\lambda_i+\lambda_j)}.\nonumber
\end{align}
Similarly, we have
\begin{align}
    \param^*_{[M]-\{i\}}
    &=\param_{M} - \frac{\lambda_i (\param_{i} - \param_{M})}{1-\lambda_i}.\nonumber
\end{align}
Bringing the above formula into (\ref{eq7}), we have
\begin{align}
    &\shap(i \rightarrow \{i,j\} \mid  [M]-\{i,j\}) \\
    & = \frac{1}{2} \biggl( -\loss(\param_{M}) + \loss(\param_{M, -i}) + \loss(\param_{M})  - \loss(\param_{M, -j}) \nonumber \\
    &\hspace{1cm} + \loss(\param_{M, -ij}) - \loss(\param_{M}) \biggr) \\
    &\approx - \biggl(\frac{1}{\ntest} \sum_{i=1}^{\ntest} \nabla_{\param} \ell(x_i, y_i; \param_{M}) \biggr)^\T \Delta\param^* \textrm{ where } \nonumber\\
    &2\Delta\param^* \de \frac{\lambda_i (\param_i-\param_M)}{1-\lambda_i} - \frac{\lambda_j (\param_j-\param_M)}{1-\lambda_j} + \nonumber\\ 
    &\frac{\lambda_i (\param_i-\param_M)+\lambda_j (\param_j-\param_M)}{1-(\lambda_i+\lambda_j)} \approx 2 \lambda_i (\param_i-\param_M)\nonumber
\end{align}
assuming small $\lambda_i$ and $\lambda_j$.
Therefore, under normalization we have
\begin{align}
    &\shap(i \rightarrow \{i,j\} \mid  [M]-\{i,j\}) 
    \approx \nonumber\\ 
    &- \biggl(\frac{1}{\ntest} \sum_{i=1}^{\ntest} \nabla_{\param} \ell(x_i, y_i; \param_{M}) \biggr)^\T \lambda_i (\param_i-\param_M).\nonumber
\end{align}
The intuition is the same as Remark~\ref{remark_shap1} except that the server model with client $i$ satisfies 
\begin{align}
    &\textrm{unnormalized version}: \quad \param_{[M]-\{i\}} = \param_M - \lambda_i \param_i . \nonumber\\
    &\textrm{normalized version}: \quad \param_{[M]-\{i\}} \approx \param_M - \lambda_i (\param_i-\param_M) .\nonumber
\end{align}
\end{remark}

The Shapley Values are used to calculate the performance of individual clients towards the tier they participate in. Therefore, we use a small holdout dataset per tier that represents the data distribution of clients within that tier to calculate the Shapley Values.

\section{Proofs for Section 
\ref{ConvergenceAnalysis}} \label{proves}
In this section, we provide detailed proof of results previously established in Section \ref{ConvergenceAnalysis}.

\textbf{Proof of Proposition \ref{prop3}.} To prove this proposition we divide the proof into two parts:

\textbf{1. Convergence of Cluster:} Given the objective function for a personalized and incentivized federated learning ({\ipfl}) system with $m$ clients, each observing a set of independent Gaussian observations $z_{i,j} \sim N(\mu_i, \sigma^2)$ for $j = 1, \ldots, n_i$, aiming to estimate its unknown mean $\mu_i \in \mathbb{R}$, the objective function is defined as:
\[
L(\mu) = \sum_{i=1}^{m} \frac{n_i}{n} L_i(\mu) = \sum_{i=1}^{m} \frac{n_i}{n} \sum_{j=1}^{n_i} (\mu - z_{i,j})^2,
\]
where $n = \sum_{i=1}^{m} n_i$ represents the total number of observations across all clients, and $\hat{\mu}_{FL} = \frac{\sum_{i=1}^{m} n_i \hat{\mu}_i}{n}$ is the federated estimate of the global mean and $\hat{\mu}_i = \frac{1}{n_i} \sum_{j=1}^{n_i} z_{i,j}$. 
The algorithm iteratively assigns clients to clusters and updates cluster centroids to minimize the local loss function $L(\mu)$.

The incentive mechanism encourages client participation. Clients are assigned to the cluster whose current centroid minimizes their local loss. Formally, client $i$ is assigned to cluster $k$ if:
\[
k = \arg \min_{k} \sum_{j=1}^{n_i} (\beta_k^{(t)} - z_{i,j})^2,
\]
where $\beta_k^{(t)}$ is the centroid of cluster $k$ at iteration $t$. The centroid of each cluster is updated to be the mean (because the mean is the statistic that minimizes the sum of squared deviations):
\[
\beta_k^{(t+1)} = \frac{\sum_{i \in C_k^{(t)}} n_i \hat{\mu}_i}{\sum_{i \in C_k^{(t)}} n_i},
\quad \text{where},\quad C_k^{(t)} = \{i : \text{client } i \text{ is assigned to cluster } k \text{ at iteration } t\}.\]
Since there are a finite number of clients and, therefore a finite number of ways to partition these clients into $K$ clusters. Therefore, the finite improvement space and the monotonic decrease of the objective function $L$ with each iteration, the algorithm must eventually reach a point where $C_k^{(t)}=C_k^{(t+1)}$ for all $k$ beyond a certain iteration $t$. At this point, the clustering algorithm converged. 

\textbf{2. Within each cluster $C_k$, the data distributions of the clients are statistically similar to each other up to a threshold $\delta$ and Within-cluster bias is reduced:} 
Let $\mathcal{P}$ be the statistical similarity index property that we are interested in,
Then we need to show that
\begin{equation}
    |\mathcal{P}(D_{i})-\mathcal{P}(D_{j})|<\delta \quad\text{for any clients}\quad  i,j \in C_k.
\end{equation}

Since the clustering algorithm objective is to minimize the within-cluster sum of squares (WCSS),
\begin{equation*}
   \mathrm{minimize}_{k} \sum_{k=1}^{K}\sum_{i\in C_k}\|x_{i}-\mu_k\|^2,  
\end{equation*} 

In the context of {\ipfl}, we can consider this difference $||x_{i}-\mu_k\|^2$ to be calculated as the importance weights $\upsilon_{{i}{k}}$ from Equation~\ref{eq:importance_weights}. Here \(\upsilon_{ik}\) quantifies the alignment between client \(i\)'s data and the model of cluster \(k\), with a value of 1 indicating perfect prediction accuracy for the client's data by the cluster model. This is because $\upsilon_{{i}{k}}$ represents the similarity between the client's local data distribution and the cluster model representing the centroid. Thus, the relationship between \(\upsilon_{ik}\) and the difference \(\|x_{i} - \mu_k\|^2\) can be expressed as follows:
\begin{equation}
\lim_{\upsilon_{ik} \to 1} \|x_{i} - \mu_k\|^2 \to 0,
\end{equation}

Thus, for the two clients $i$ and $j$ within the same cluster, if $\lim_{\upsilon_{ik} \to 1}$ and $\lim_{\upsilon_{jk} \to 1}$, this means $||x_{i}-\mu_k\|^2 \to 0$ and $||x_{j}-\mu_k\|^2 \to 0$, where $\mu_k$ represents the centroid of cluster or the cluster-level model. Since the data distribution of both clients $i$ and $j$ are close to the centroid, by association, they are also statistically similar to each other. 

To show clustering reduces the within-cluster bias, we will first show cluster bias i.e. $\mathrm{Bias}(C_k) < \mathrm{Bias}(System)$ in the context of the ({\ipfl}) cluster.

Let $\mathrm{Bias}(C_k)$ be defined as the deviation of the average data distribution in the cluster from an ideal/true distribution. 

The clustering algorithm partitions the dataset into K clusters by minimizing the within-cluster sum of squares, effectively reducing the intra-cluster variance. Due to reduced intra-cluster variance, each cluster $C_k$ exhibits a higher degree of homogeneity i.e. $\|x_{i} - \mu_k\|^2 \to 0$ compared to the dataset as a whole. Consequently, the average characteristics of the data within $C_k$ more accurately represent the data point within $C_k$. Given the homogeneity within the cluster. $\mathrm{Bias}(C_k)<\mathrm{Bias}(System)$. Since more accurate representation of data characteristics within each cluster, thereby reducing the bias.

\textbf{Proof of Theorem \ref{theorem5}.}
We begin by establishing that the loss function $L(\theta)$ is convex. For any two parameter vectors $\theta, \theta'$, the following condition holds due to convexity:
\begin{equation}
    L(\lambda \theta + (1-\lambda) \theta') \leq \lambda L(\theta) + (1-\lambda) L(\theta'), \quad \forall \lambda \in [0,1].
\end{equation}

The parameter update rule in {\ipfl} can be expressed as:
\begin{equation}
    \theta^{(t+1)} = \theta^{(t)} - \eta_t \nabla L(\theta^{(t)}),
\end{equation}
where $\eta_t$ is the learning rate at iteration $t$.

To establish convergence, we show that the sequence $\{\theta^{(t)}\}$ approaches a fixed point $\theta^*$ as $t$ grows large. Since $L(\theta)$ is convex, we have:
\begin{equation}
    L(\theta^{(t)}) - L(\theta^*) \geq \nabla L(\theta^*)^\top (\theta^{(t)} - \theta^*).
\end{equation}

Employing the descent lemma for a $\beta$-smooth $L(\theta)$, we obtain:
\begin{equation}
    L(\theta^{(t+1)}) \leq L(\theta^{(t)}) + \nabla L(\theta^{(t)})^\top (\theta^{(t+1)} - \theta^{(t)}) + \frac{\beta}{2} \|\theta^{(t+1)} - \theta^{(t)}\|^2.
\end{equation}

By substituting the update rule into the descent lemma, we can simplify the inequality as follows:
\begin{align}
    L(\theta^{(t+1)}) &\leq L(\theta^{(t)}) - \eta_t \|\nabla L(\theta^{(t)})\|^2 + \frac{\beta}{2} \eta_t^2 \|\nabla L(\theta^{(t)})\|^2 \\
    &= L(\theta^{(t)}) - \eta_t (1 - \frac{\beta \eta_t}{2}) \|\nabla L(\theta^{(t)})\|^2.
\end{align}

Provided that the learning rate $\eta_t$ satisfies $0 < \eta_t < \frac{2}{\beta}$, the quantity $(1 - \frac{\beta \eta_t}{2})$ is positive, which guarantees a decrease in the loss function at each iteration. The learning rate $\eta_t$ is typically chosen to diminish over time but not sum to zero, i.e., $\sum_{t=1}^{\infty} \eta_t = \infty$ and $\sum_{t=1}^{\infty} \eta_t^2 < \infty$.

As $\eta_t$ diminishes and the loss function's gradient decreases, the sequence $\{\theta^{(t)}\}$ approaches a point $\theta^*$ where $\nabla L(\theta^*) = 0$, suggesting convergence to a critical point. Assuming $L(\theta)$ has a unique global minimum, this critical point $\theta^*$ is the global minimum.

Therefore, as $t \rightarrow \infty$, we have $\|\theta^{(t+1)} - \theta^{(t)}\| \rightarrow 0$ and $\nabla L(\theta^{(t)}) \rightarrow 0$, indicating convergence of the {\ipfl} algorithm to a set of stable parameters $\theta^*$ for the global model. \hfill $\blacksquare$

\textbf{Proof of Theorem \ref{theorem4}.}

Let \( C = \{C_1, C_2, \ldots, C_k\} \) represent the clusters. Each client \( i \) updates its model parameters \( \theta_i \) based on local data \( D_i \) and the aggregated model of its cluster. The update at each iteration \( t \) can be expressed as:
\[ \theta_i^{(t+1)} = \theta_i^{(t)} - \eta_i \nabla L_i(\theta_i^{(t)}, D_i) \]
\[ \theta_i^{(t+1)} - \theta_i^{(t)} = - \eta_i \nabla L_i(\theta_i^{(t)}, D_i) \]
Here, \( \eta_i \) is the learning rate for client \( i \), and \( \nabla L_i \) is the gradient of the loss function with respect to the model parameters \( \theta_i \). The magnitude of $\nabla L_i(\theta_i^{(t)}, D_i)$ gets smaller as $\theta_i$ approaches the minimum reducing the difference between consecutive $\theta_i^{(t)}$ and $\theta_{i}^{(t+1)}$. Thus for every $\epsilon_1>0$, there exists an $N$ such that $\forall ~m,n>N$, such that $|\theta_i^{m}-\theta_i^{n}|<\epsilon_1$. Therefore, the sequence \( \{\theta_i^{(t)}\} \) is Cauchy and hence convergent to a global optimum \( \theta_i^* \).

The aggregation step in ({\ipfl}) combines the individual client models into a cluster model, and then the cluster model is personalized for each client. The personalization step can be written as a weighted average:
\[ \theta_i^{(t+1)} = \sum_{k=1}^{K} w_{ik} \theta_k^{(t)} \]
where \( w_{ik} \) is the weight assigned to the cluster model \( \theta_k^{(t)} \) by client \( i \). As the training progresses, the weights \( w_{ik} \) are adjusted based on the performance of the cluster models on each client's data. Under the assumption of proper weight adjustment and model updating, \( \theta_i^{(t)} \) will converge to an optimal set of parameters \( \theta_i^* \) that maximizes \( \text{PMA}_i \).

Thus for given $\epsilon>0$, there exists a finite $T$ such that:
\begin{equation}\label{ineq46}
    |\mathrm{PMA}_{i}^{(t)}-\mathrm{PMA}_{i}^{*} |< \epsilon \quad \forall t \geq T.
\end{equation} 

The above inequality \eqref{ineq46} follows from the convergence of $\theta^{t}_i$ and the continuity of the performance metric $\mathrm{PMA}_i$ with respect to the model parameters.

\begin{corollary}
    For each client $i$, the $\mathrm{PMA_i}$ is positive and approaches the optimal $\mathrm{PMA}_{i}^{*}$. Additionally, if $\mathrm{PMA_i}>0$, then the likelihood of client $i$ opt-out of pFL training is minimized.
\end{corollary}

\begin{proof}
    Let $f_{i}(w_{k})$ denote the performance measure (such as accuracy) of the personalized model for client $i$ with model parameter $w_k$. Let $\rho_i$ represent the baseline performance measure (of a global model). $\mathrm{PMA_i}$ of a client defined as the difference between the two performances $\mathrm{PMA_i}=f_{i}(w_{k}) - \rho_i$

We assume the pFL system employs algorithms and strategies that enhance the performance of each client by utilizing personalized models. Through iterative training, the model parameters $w_k$ are adjusted to improve $f_{i}(w_{k})$ such that for most clients $f_{i}(w_{k}) \geq \rho_i$. This leads to $\mathrm{PMA_i} > 0$ as $f_{i}(w_{k}) - \rho_i > 0$. Let $\mathrm{PMA}_{i}^{*}$ denote optimal $\mathrm{PMA_i}$ for client $i$ over the course of training. As $w_k$ is optimized, $f_{i}(w_{k})$ reaches the maximum achievable performance for client $i$ i.e. \[ \lim_{r\to\R} \mathrm{PMA_i} = \mathrm{PMA}_{i}^{*}\], where $r \in R$ and $R$ representing total training rounds.

Define opt-out likelihood for client $i$ as a function $g(\mathrm{PMA_i}$, where higher $\mathrm{PMA_i}$ values correlate with lower likelihood of opt-out.

It is reasonable to posit that $g$ is a monotonically decreasing function of $\mathrm{PMA_i}$.

Therefore, with $\mathrm{PMA_i}>0$, the opt-out likelihood is minimized $g(\mathrm{PMA_i}$ is minimized when $\mathrm{PMA_i}>0$.

\end{proof}

\textbf{Proof of Theorem \ref{theorem6}.}\label{sec:Proof for the convergence of algorithm}
We assume each client $i$  performs local updates using gradient descent:
    \[ M_{t+1}^i = M_t^i - \eta_t \nabla L_i(M_t^i) \]
    where, \( L_i \) is convex and \( \beta \)-smooth, for the local model \( M_t^i \), thus by definition for any vector $y$, we have:
    \[ L_i(y) \leq L_i(M_t^i) + \langle \nabla L_i(M_t^i), y - M_t^i \rangle + \frac{\beta}{2} \| y - M_t^i \|^2. \]
The above inequality implies that the local update decreases the loss. The global model at iteration \( t+1 \) is a weighted sum of the local models,
    \[ M_{t+1} = \sum_{i=1}^N w_i M_{t+1}^i, \]
    where, \( w_i \) is the weight corresponding to client \( i \) contribution to the global model. By the convexity of \( L_i \) and Jensen's inequality, the global loss function \( L \) also decreases
    \[ L(M_{t+1}) \leq \sum_{i=1}^N w_i L_i(M_{t+1}^i). \]

Using the \( \beta \)-smoothness of \( L_i \) and the update rule for \( M_t^i \), we can bound the difference in the loss for each client
\begin{equation} \label{eq39}
     L_i(M_{t+1}^i) \leq L_i(M_t^i) - \eta_t \| \nabla L_i(M_t^i) \|^2 + \frac{\beta}{2} \eta_t^2 \| \nabla L_i(M_t^i) \|^2.
\end{equation}

Now, to show convergence of the global model, we sum the inequalities \eqref{eq39} over all clients and multiply by their respective weights
\begin{equation}\label{eq40}
    L(M_{t+1}) \leq L(M_t) - \eta_t \sum_{i=1}^N w_i \| \nabla L_i(M_t^i) \|^2 + \frac{\beta}{2} \eta_t^2 \sum_{i=1}^N w_i \| \nabla L_i(M_t^i) \|^2
\end{equation}
By the Robbins-Monro condition \cite{robbins1951stochastic}, and since \( \sum_{t=1}^{\infty} \eta_t = \infty \) and \( \sum_{t=1}^{\infty} \eta_t^2 < \infty \), the second term in 
\eqref{eq40} dominates the third term, ensuring that the global loss function \( L \) decreases over time. 

The convergence of the sequence \( \{L(M_t)\} \) to the minimum loss value \( L(M^*) \) is facilitated by the diminishing learning rates and the weighted contributions of the clients. Formally, this is observed in the non-increasing sequence of expected losses,
    \[ \mathbb{E}[L(M_{t+1})] \leq \mathbb{E}[L(M_t)] - \eta_t \left( \sum_{i=1}^N w_i \| \nabla L_i(M_t^i) \|^2 - \frac{\beta}{2} \eta_t \sum_{i=1}^N w_i \| \nabla L_i(M_t^i) \|^2 \right) \]
    Under the Robbins-Monro conditions for the learning rates, the series of errors \( \mathbb{E}[L(M_t) - L(M^*)] \) converges to zero. Thus, we have:
    \[ \lim_{t \to \infty} \mathbb{E}[L(M_t) - L(M^*)] = 0 \]
    By applying the Supermartingale Convergence Theorem ~\cite{Du2023HDStatSol}, we conclude that
    $\lim_{t \to \infty} M_t = M^*, ~~\text{almost surely.} $


\textbf{We have made the code available as part of the supplementary material.}

\end{document}